\documentclass{article}

\PassOptionsToPackage{numbers}{natbib}

\usepackage[preprint]{neurips_2024}

\usepackage[utf8]{inputenc} 
\usepackage{titlesec}

\usepackage{tocloft}

\cftsetindents{section}{0em}{2em}
\cftsetindents{subsection}{0em}{2em}
\usepackage[T1]{fontenc}    
\usepackage[colorlinks=true, linkcolor=black, citecolor=blue, urlcolor=blue]{hyperref}
\usepackage{url}            
\usepackage{booktabs}       
\usepackage{amsfonts}       
\usepackage{nicefrac}       
\usepackage{microtype}      
\usepackage{wrapfig}
\usepackage{graphicx} 
\usepackage{tabularx}
\usepackage{cancel}
\usepackage{bbding}
\usepackage{amsmath}
\usepackage{enumitem}
\usepackage{multicol}
\usepackage{multirow}
\usepackage{makecell}
\usepackage{balance}
\usepackage{xspace}
\usepackage{colortbl}
\usepackage{ulem}
\usepackage{subfig}
\usepackage{graphbox}
\usepackage{svg}
\usepackage{nicefrac}
\usepackage{soul}
\usepackage{bold-extra}
\usepackage{array}
\newcolumntype{P}[1]{>{\centering\arraybackslash}p{#1}}
\newcolumntype{M}[1]{>{\centering\arraybackslash}m{#1}}
\newcolumntype{R}[1]{>{\arraybackslash}m{#1}}

\title{Towards Trustworthy Retrieval Augmented Generation for Large Language Models: A Survey}
\author{%
Bo Ni$^{1}$, 
Zheyuan Liu\textsuperscript{\textdagger}$^{2}$,
Leyao Wang\textsuperscript{\textdagger}$^{1}$
Yongjia Lei\textsuperscript{\textdagger}$^{3}$, 
Yuying Zhao$^{1}$, 
Xueqi Cheng$^{1}$,\\
\textbf{Qingkai Zeng$^{2}$,
Luna Dong$^{4}$, 
Yinglong Xia$^{4}$, 
Krishnaram Kenthapadi$^{5}$, 
Ryan Rossi$^{6}$,} \\
\textbf{Franck Dernoncourt$^{6}$, 
Md Mehrab Tanjim$^{6}$, 
Nesreen Ahmed$^{7}$, 
Xiaorui Liu$^{8}$, 
Wenqi Fan$^{9}$,} \\
\textbf{Erik Blasch$^{10}$, 
Yu Wang\textsuperscript{*}$^{3}$, 
Meng Jiang\textsuperscript{*}$^{2}$, 
Tyler Derr\textsuperscript{*}$^{1}$}\\
\\
$^1$Vanderbilt University,  
$^2$University of Notre Dame, 
$^3$University of Oregon, 
$^4$Meta,\\ 
$^5$Oracle Health AI, 
$^6$Adobe Research, 
$^7$Cisco AI Research, 
$^8$North Carolina State University,\\ 
$^9$The Hong Kong Polytechnic University,  
$^{10}$Air Force Research Lab\\
\\
  \texttt{\{bo.ni, leyao.wang, yuying.zhao, xueqi.cheng, tyler.derr\}@vanderbilt.edu}, \\
  \texttt{\{zliu29, qzeng, mjiang2\}@nd.edu}, 
  \texttt{\{yongjia, yuwang\}@uoregon.edu},\\ 
  \texttt{\{lunadong, yxia\}@meta.com}, 
  \texttt{krishnaram.kenthapadi@oracle.com}, \\
  \texttt{\{ryrossi,dernonco,tanjim\}@adobe.com}, 
    \texttt{nesahmed@cisco.com}, \\ \texttt{xliu96@ncsu.edu}, 
    \texttt{wenqi.fan@polyu.edu.hk}, \texttt{erik.blasch.1@us.af.mil}
}

\begin{document}

\maketitle

\renewcommand{\thefootnote}{\fnsymbol{footnote}}
\footnotetext[0]{\textsuperscript{\textdagger}Significant Contribution.}
\footnotetext[0]{\textsuperscript{*}Corresponding Authors.}

\begin{abstract}
\vspace{2ex}
Retrieval-Augmented Generation (RAG) is an advanced technique designed to address the challenges of Artificial Intelligence-Generated Content (AIGC). By integrating context retrieval into content generation, RAG provides reliable and up-to-date external knowledge, reduces hallucinations, and ensures relevant context across a wide range of tasks. However, despite RAG's success and potential, recent studies have shown
that the RAG paradigm also introduces new risks, including robustness issues,
privacy concerns, adversarial attacks, and accountability issues. Addressing
these risks is critical for future applications of RAG systems, as they
directly impact their trustworthiness. Although various methods have been
developed to improve the trustworthiness of RAG methods, there is a lack of a unified
perspective and framework for research in this topic. Thus, in this paper, we
aim to address this gap by providing a comprehensive roadmap for developing
trustworthy RAG systems. We place our discussion around five key perspectives: reliability, privacy, safety, fairness, explainability, and accountability. For each perspective, we present a general framework and taxonomy, offering a structured approach to understanding the current challenges, evaluating existing solutions, and identifying promising future research directions. To encourage broader adoption and innovation, we also highlight the downstream applications where trustworthy RAG systems have a significant impact. For more information about the survey, please check our GitHub repository\footnote{\url{https://github.com/Arstanley/Awesome-Trustworthy-Retrieval-Augmented-Generation}}.

\end{abstract}

\maketitle
\newpage 

\tableofcontents

\section{Introduction}
Retrieval-Augmented Generation (RAG) has emerged as a promising approach to
address the challenges faced by Large Language Models (LLMs), such as
hallucinations, reliance on outdated knowledge, and the lack of explainability~\cite{Gao2023Retrieval, Zhang2023Sirens}. By incorporating external information into the generation context, RAG improves the accuracy and reliability of the generated content. The recency of information also enables the model to stay current with minimal training costs by reducing the need for extensive retraining of the entire system to update its parameters. These benefits have profound implications for
real-world applications. For example, RAG has been effectively applied in
medical question answering~\cite{xiong2024benchmarkingretrievalaugmentedgenerationmedicine, Zakka2023Almanac, singhal2023expertlevelmedicalquestionanswering}, legal
document drafting~\cite{Wiratunga2023CBR, pipitone2024legalbenchragbenchmarkretrievalaugmentedgeneration}, educational
chatbots~\cite{Thway2023Battling}, and financial report summarization~\cite{yepes2024financialreportchunkingeffective} due to their adaptability in various domains. 

The definition of trustworthiness often depends on the context of discussion~\cite{trustworthy_ai_computational_perspective_2021, survey_trustworthy_ai_meta_decision_2023, trustworthy_ai_acm_2023, global_study_trust_ai_2023, trustworthy_gnn_2023, liu2024machine, trustworthy_graph_neural_networks_2024, trustworthy_llms_2024}. In the context of machine learning and artificial intelligence, trustworthy AI must exhibit characteristics that make the system \textit{worthy of trust}. In 2022, the National Institute of Standards and Technology (NIST) published guidelines for trustworthy AI, defining trustworthiness from several perspectives~\cite{nist_trustworthy_ai}: Reliability, Privacy, Explainability, Fairness, Accountability, and Safety. 

\textbf{Reliability} ensures that the system consistently performs as expected and produces accurate results under various conditions. It includes addressing challenges such as uncertainty quantification and robust generalization, which are critical for enhancing system dependability. For instance, in a legal case analysis system, reliability involves balancing uncertainty quantification (e.g., confidence scores for retrieved legal citations and the number of retrieved legal citations) and robust generalization (e.g., applying precedents to new cases) to ensure lawyers are not misled during case preparation.

\textbf{Privacy} focuses on safeguarding user data, ensuring control over personal information. Since RAG has been applied to sensitive domains like the medical field, protecting patient information is important. For example, when a healthcare assistant retrieves medical records or generates treatment suggestions, the system must prevent data breaches and ensure sensitive patient details embedded in the language model remain secure. 

\textbf{Explainability} emphasizes the need for transparent decision-making processes, enabling users to understand how outputs are generated. For example, a university admissions assistant powered by RAG should offer clear explanations of how student profiles are matched with program requirements, providing insights that users can readily understand and verify.

\textbf{Fairness} focuses on minimizing biases introduced during both retrieval and generation stages, as these biases can significantly affect outcomes in high-stakes domains. Recent advancements include the use of re-ranking methods to mitigate societal biases in retrieval and fine-tuning techniques to balance demographic fairness with system performance. 
For example, the admissions assistant must ensure fair treatment of applicants by addressing potential biases, such as those related to gender, race, or socioeconomic status.

\textbf{Accountability} pertains to the governance of AI, including policymaking and law enactment, but also extends to technical aspects such as tracing the origins and processes behind AI-generated content. For example, ensuring that a news-generating system can trace its retrieved sources to improve content accountability and reduce misinformation is critical. Techniques like content watermarking help identify the provenance of retrieved information and the generation process, providing a clear audit trail for future verification.

\textbf{Safety} addresses the system's capacity to prevent and mitigate harm, with a particular focus on defending against adversarial attacks and reducing risks from malicious actors. Current chatbot systems often interact with high-risk users, such as teenagers, who may unknowingly be exposed to harmful or inappropriate content. Adversarial attacks and jailbreaking attempts that alter the chatbot's behavior could lead to misinformation, inappropriate responses, or even dangerous suggestions. Thus, building robust safeguards, such as adversarial training and ethical guardrails, is crucial for ensuring safety and preventing harm in such interactions.

Despite their recent success, concerns about the trustworthiness of RAG-based
systems have become an increasing subject of debate. First, RAG systems are susceptible to reliability
issues since developers must ensure the output is accurately grounded on the retrieved content~\cite{li2023traq, Gao2023Retrieval}. Second, the reliance on an external database introduces a new attack surface, exposing the systems to a range of adversarial threats~\cite{dual_rag_2024, adaptive_adversarial_rag_2024,
robust_rag_2024, robust_rag_iclr_2024, misinformation_rag_2024,
alignment_rag_2023}. As a result, robustness improvements are needed to
safeguard the systems. Third, RAG systems pose new challenges regarding data
privacy~\cite{privacy_rag_2024}. The integration of external databases introduces additional leakage channels. It is imperative to ensure that the RAG systems do
not expose private information from both the external databases and the training data of the underlying LLM during the generation process. Additionally, RAG could be
susceptible to fairness issues~\cite{shrestha2024fairrag} from both the retrieving process and the generation process. How the retrieved data is
selected and utilized can significantly affect the fairness of the generated
content. The implicit bias during the generation could also be affected by the retrieved content due to the increased confidence~\cite{no_free_lunch_rag_2025}. Lastly, with the rise and potential use of LLMs, accountability is a subject for policymakers on the use of RAG systems. Although progress has been made, these
challenges significantly restrict the wide adoption of RAG systems in real-world
scenarios, especially in high-stakes scenarios such as medication, legal
consulting, and education~\cite{xiong2024benchmarkingretrievalaugmentedgenerationmedicine, Zakka2023Almanac,
Wiratunga2023CBR, Thway2023Battling}. Thus, it is essential to incorporate the
trustworthy perspective while advancing the RAG systems. 

Due to the importance of trustworthy AI, a plethora of research has been
developed to advance the application of RAG in Large Language Models with heterogeneous definitions, tremendous
implementation, and inconsistent evaluation metrics. However, there is no
systematic review of this area's current advancements and challenges. To organize the various perspectives, this paper formulates a systematic discussion on the state of
trustworthy RAG in Large Language Models. 
The list of papers discussed is provided in the GitHub repository\footnote{\url{https://github.com/Arstanley/Awesome-Trustworthy-Retrieval-Augmented-Generation}}.

\begin{figure*}[t]
 
    \centering
    \includegraphics[width=0.9\textwidth]{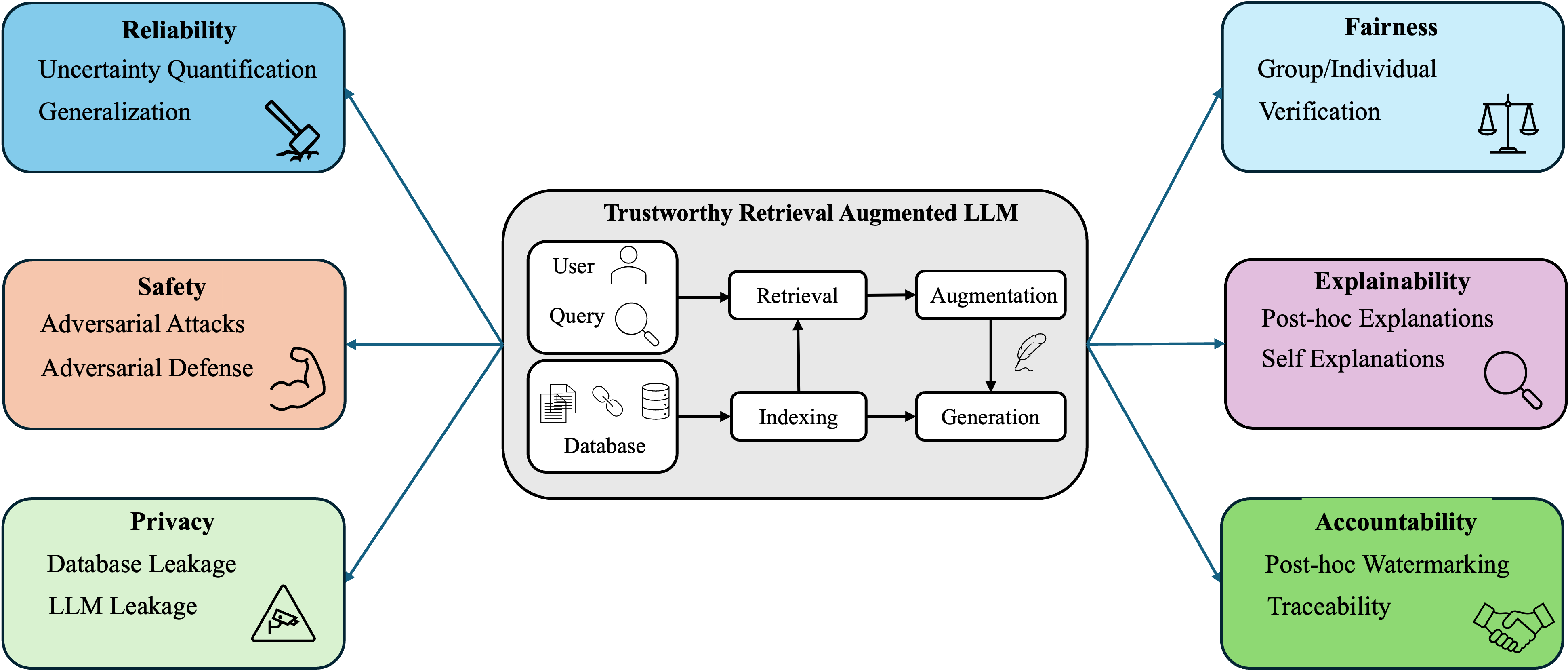}
    \label{fig:overview}
    \caption{An overview of the key components and dimensions of Trustworthy Retrieval Augmented Generation (RAG) for Large Language Models (LLMs) that are covered in this survey.  }
    \vspace{-0.1in}
\end{figure*}

\section{Preliminaries}
This section provides the preliminaries of the RAG framework for LLMs. We will introduce the concept of RAG and the common downstream tasks. We acknowledge the wide range of applications of RAG in domains other than LLMs (e.g., Image Generation), but this survey limits the scope to the applications of RAG in LLMs, sometimes referred to as Retrieval Augmented Language Models (RA-LLMs)~\cite{fan2024survey}. As a simplification of the terminology, in the rest of this survey, we use RAG, RA-LLM, and RAG-based systems interchangeably. 

\subsection{Retrieval Augmented Generation}
As illustrated in Figure \ref{fig:overview}, a typical RAG framework consists of three stages: information retrieval, knowledge augmentation, and content generation. Given a query, the retrieval process aims to provide relevant information and context to facilitate the reasoning of the query. Following the classification of previous work~\cite{Gao2023Retrieval}, the retrieval process contains two stages - \textit{indexing} and \textit{retrieving}. The \textit{indexing} stage takes inputs from a diverse range of formats (PDFs, HTML, words, Markdowns, etc.) and converts them into chunks of data. Subsequently, the chunks of data are converted into vector representations and stored in a vector database for access during inference. However, it is worth noting that for some retrieval-augmented tasks, such as those involving knowledge graphs (e.g., GraphRAG frameworks), the external databases may not rely on vector databases. Instead, they often use symbolic knowledge structures or relational data to store and retrieve information, bypassing the need for vectorization. The \textit{retrieval} stage takes the incoming query, converts it into vector representation (or aligns it with the symbolic knowledge structures in non-vector cases), and then fetches the matching chunks or nodes from the database. \textit{Knowledge augmentation} involves integrating the retrieved knowledge into the underlying generation models. Often it is achieved through injecting the knowledge into the prompt or finetuning the language model on the retrieved knowledge. Finally, the \textit{generation} stage utilizes the augmented knowledge to produce coherent and contextually accurate responses based on the query.  

Compared to the direct generation by LLMs without any contextual information retrieved externally, the additional consideration of external knowledge in RAG paradigm leads to new challenges in trustworthiness. Each stage - retrieval, augmentation, and generation - introduces unique trustworthy challenges, such as retrieval bias, hallucination, and the injection of irrelevant information, all of which require careful mitigation to ensure a trusted response. These concerns will be further explored in subsequent sections of this survey.

\subsection{Tasks and Evaluations}
\subsubsection{Tasks}
\label{sec:eval}
The RAG paradigm has enabled multiple applications in the natural language processing (NLP) domains. The following briefly introduces some of the common tasks including question answering and chatbots, while also introducing some of the associated and commonly used datasets.

\paragraph{Question Answering.}  
One of the primary downstream tasks for RAG-based language models is question
answering (QA), which includes various sub-tasks such as long-form question
answering, multi-hop question answering, domain-specific question answering, and
open-domain question answering. In these tasks, the system is given a user query
and aims to generate the most relevant and accurate answer. The RAG paradigm assists these tasks by integrating the relevant external context into the generation.

Different QA sub-tasks might require different evaluation paradigms. For example, for multi-hop and multiple choice question answering, since the answer is relatively structured or limited to a pre-defined set of answers, the performance is often evaluated through metrics such as hits@$n$, where $n$ represents the rank of the correct answer in the list of retrieved candidates, and F1, where precision and recall are balanced to assess both the correctness of the generated answers and the system's ability to retrieve all relevant answers~\cite{luo2024rog, sun2023thinkongraph}. On the other hand, for open domain question answering where answers are more unstructured, Exact Match (EM) and Lexical Match are commonly used~\cite{zhang-etal-2023-survey-efficient}. These metrics both provide a good measure of quality of the generated answers. It is also worth mentioning that for some works that emphasize the retrieval process also report results that measures the quality of the retrieval. Commonly used QA datasets include the following:

\begin{itemize}
    \item \textbf{MMLU}~\cite{hendrycks2021measuringmassivemultitasklanguage}: A commonly used dataset for multiple-choice question answering (MCQA). It contains MCQA questions from 57 domains, including STEM (science, technology, engineering, and math), humanities, and medicine. In their experiments, existing uncertainty quantification research chooses to use a subset of the dataset for evaluation~\cite{ye2024benchmarkingllmsuncertaintyquantification, kumar2023conformalpredictionlargelanguage}.
    
    \item \textbf{TriviaQA}~\cite{joshi2017triviaqalargescaledistantly}: A widely-used large-scale dataset for open-domain question answering, designed to test models on questions from Wikipedia and web search engines. TriviaQA includes question-answer pairs along with evidence documents for context, making it suitable for testing reading comprehension and retrieval-based models. It is used by most open-domain question answering models~\cite{su2024apienoughconformalprediction, quach2024conformallanguagemodeling, li2023traq}.

    \item \textbf{WebQSP}~\cite{yih2016value}: A popular dataset for multi-hop knowledge base question answering, focusing on the task of answering questions by traversing multiple entities and relations within a knowledge graph. WebQSP provides questions labeled with their corresponding semantic parses, enabling models to learn complex query structures for effective knowledge graph traversal and reasoning~\cite{ni2024trustworthyknowledgegraphreasoning, luo2024rog, sun2023thinkongraph}. 
\end{itemize}

\paragraph{Chatbots.} Another common application for RAG-based language models
is Chatbots. Chatbots are designed to handle an array of dialogue types, ranging
from task-oriented interactions to open-domain conversations. The RAG paradigm
can enhance Chatbots performance by integrating external knowledge into the
conversation, allowing the
system to access up-to-date information that would otherwise be out of its
parameters trained from outdated text corpus~\cite{akkiraju2024factsbuildingretrievalaugmented, sumida2024ragchatbotsforgetunimportant, feldman2024raggededgesdoubleedgedsword}. Contemporary knowledge will be especially important
for those Chatbots operating in dynamic or domain-specific environments, such as
medical and financial dialogue systems.

When evaluating the performance of a chatbot, two primary goals are considered. The first goal is to assess the quality and coherence of the dialogue. To achieve this, various metrics have been proposed, including those that measure utility~\cite{Cameron2019}, response understanding~\cite{Yokotani2018}, and overall aesthetics~\cite{Wargnier2018}. Additionally, some metrics evaluate the similarity between the generated responses and human responses, as seen in works such as~\cite{Adiwardana2020, Xu2022}. While these metrics are generally effective for open-domain dialogues, they often fail to account for specific use cases. Thus, the second goal focuses on evaluating the chatbot's effectiveness in meeting users' needs, particularly in real-world, business-oriented use cases. In these scenarios, beyond the aforementioned dialogue quality metrics, practical effectiveness is a key consideration. General natural language generation (NLG) metrics, such as BLEU~\cite{10.3115/1073083.1073135} and ROUGE~\cite{lin2004rouge}, are frequently employed to measure this aspect.

There is currently no standardized holistic evaluation of chatbots. For large commercial models, the performance of the chatbots is evaluated on the subtasks such as code generation, problem solving, and complex reasoning. On the other hand, for the business-oriented, task specific models, they are evaluated on the corresponding task-specific or industry-specific datasets. We introduce some of the commonly used datasets below. 

\begin{itemize}
    \item \textbf{HellaSwag}~\cite{zellers2019hellaswag}: A commonly used dataset for commonsense reasoning and story completion tasks. HellaSwag presents models with scenarios requiring contextually appropriate completions, testing their ability to reason beyond surface-level semantics. It has been widely adopted for benchmarking commonsense reasoning capabilities in large language models~\cite{openai2024gpt4technicalreport, brown2020languagemodelsfewshotlearners}.
    
    \item \textbf{HumanEval}~\cite{chen2021evaluatinglargelanguagemodels}: A widely used dataset for evaluating code generation capabilities of language models. HumanEval includes programming problems of varying difficulty levels, along with unit tests to validate the correctness of generated code. It is a standard benchmark for assessing the coding performance of generative models~\cite{openai2024gpt4technicalreport}.
    
    \item \textbf{MedicationQA}~\cite{benabacha2019bridging}: A popular dataset for question answering in the medical domain, focusing on patient-generated questions about medication. It includes complex medical queries paired with evidence-based answers, making it a crucial benchmark for evaluating the applicability of language models in healthcare and patient communication~\cite{openai2024gpt4technicalreport, lee2023gpt4medicine}.
\end{itemize}

\paragraph{Others.}
Beyond language-based tasks, RAG-based language models can be applied to a diverse range of downstream tasks, including recommendation systems~\cite{li2023gpt4recgenerativeframeworkpersonalized}, software engineering~\cite{fan2023largelanguagemodelssoftware}, and AI for scientific discovery~\cite{ai4science2023impactlargelanguagemodels}. However, these applications are often overlooked in discussions of trustworthy RAG frameworks. Recognizing their importance, we highlight the need for further exploration of trustworthiness in these domains and propose to address them in future directions.

\paragraph{Trustworthy Evaluation.} To comprehensively assess trustworthiness, additional metrics are necessary, including those that evaluate bias, fairness, and reliability in the generated answers. As AI trust evaluation has well been debated across cognitive, communication, information, and social dimensions, the focus was on the use of the result. To maintain trust, a trust worthy evaluation is also needed to determine the efficacy under changing conditions. These metrics will be designed to ensure that the system aligns well with trustworthy aspects. Due to their heterogeneous nature, we will delve into these specific metrics in detail in the corresponding sections throughout the rest of the survey.

\subsection{Motivation}

Although trustworthiness in deep learning and LLMs has been well-explored in the general AI community, it is rather critical to consider the trustworthiness for RAG-based LLMs because (a) their growing usage in real-world applications often involves high-stakes decision-making, where errors or biases can lead to significant consequences; (b) RAG models are vulnerable to trustworthiness issues as the multiple stages in RAG—such as indexing, retrieval, and generation—can introduce compounding errors, biases, or hallucinations, which can be difficult to trace and mitigate; and (c) RAG-based LLMs are inherently different from traditional standalone LLMs as they rely on external data sources during the retrieval process, leading to a dynamic interaction between the model and potentially unreliable or biased external information. Thus, directly extending the trustworthiness framework for standalone LLMs to RAG-based LLMs is challenging due to evolving complexities. Therefore, developing a comprehensive and robust trustworthiness framework specifically tailored for RAG-based LLMs is imperative to ensure reliable and safe deployment in various domains. 

While prior work~\cite{privacy_rag_2024, xue2024badrag, robust_rag_2024, doshi2017accountability, zhou2024trustworthiness} has touched on these issues individually, there remains a gap in the literature for a unified survey that reviews the current advancements and challenges in ensuring trustworthiness for RAG-based LLMs. This survey aims to be a comprehensive review by summarizing existing efforts, categorizing approaches, and identifying opportunities for future research. By providing a systematic review, we hope to facilitate further research and development of trustworthy RAG-based LLMs across diverse domains.

\begin{table}[]
\label{survey-taxonomy}
\centering 
\scriptsize 
\setlength\tabcolsep{4pt}
\caption{Comparison with Existing Surveys on RAG and Trustworthy LLMs. (We mark some of the items 
\bcancel{\Checkmark} for LLM related survey as they do not focus on RAG) }

\begin{tabular}{ll|cccccccccccc}
\toprule
  &\multicolumn{1}{c|}{}  & \multicolumn{12}{c}{\textbf{Pillars of Trustworthiness}} 
  \\ \cmidrule(l){3-14} 
    & & \multicolumn{2}{c|}{\textbf{Reliability}}                                                                                                                                                    & \multicolumn{2}{c|}{\textbf{Privacy}}                                                                                                                                                  & \multicolumn{2}{c|}{\textbf{Safety}}                                                                                                                   & \multicolumn{2}{c|}{\textbf{Fairness}}                                                                                                                     & \multicolumn{2}{c|}{\textbf{Explainability}} & \multicolumn{2}{c}{\textbf{Accountability}}                                                                                           \\  \cmidrule(l){3-14} 
                                                          
\multicolumn{2}{c|}{\textbf{Surveys}} & \rotatebox{60}{\textbf{Uncertainty}} & \multicolumn{1}{c|}{\rotatebox{60}{\textbf{Generalizability}}} & \rotatebox{60}{\textbf{External}} & \multicolumn{1}{c|}{\rotatebox{60}{\textbf{Training Data}}} & \rotatebox{60}{\textbf{Jailbreaking}} & \multicolumn{1}{c|}{\rotatebox{60}{\textbf{Defense}}} & \rotatebox{60}{\textbf{Retrieval}} & \multicolumn{1}{c|}{\rotatebox{60}{\textbf{Generation}}} & \rotatebox{60}{\textbf{Retrieval}} & \multicolumn{1}{c|}{\rotatebox{60}{\textbf{Generation}}} & \rotatebox{60}{\textbf{Retrieval}} & \multicolumn{1}{c}{\rotatebox{60}{\textbf{Generation}}}\\ \midrule

\multirow{2}{*}{LLM}& ~\cite{trustworthy_llms_2024}     &    \bcancel{\Checkmark}  &  \bcancel{\Checkmark} & \XSolidBrush & \bcancel{\Checkmark} & \bcancel{\Checkmark} & \bcancel{\Checkmark} & \XSolidBrush & \bcancel{\Checkmark}  & \XSolidBrush & \bcancel{\Checkmark}  & \XSolidBrush &    \bcancel{\Checkmark}                                                               \\ & ~\cite{huang2023surveysafetytrustworthinesslarge}     &   \XSolidBrush  &  \XSolidBrush & \XSolidBrush & \bcancel{\Checkmark} & \bcancel{\Checkmark}  & \bcancel{\Checkmark} & \XSolidBrush &      \bcancel{\Checkmark}                                     & \XSolidBrush & \Checkmark     & \XSolidBrush & \bcancel{\Checkmark} \\ \midrule 
& ~\cite{fan2024survey} & \XSolidBrush & \XSolidBrush & \XSolidBrush & \XSolidBrush & \XSolidBrush & \XSolidBrush & \XSolidBrush & \XSolidBrush & \Checkmark & \XSolidBrush  & \XSolidBrush & \XSolidBrush\\
\multirow{3}{*}{RAG}& ~\cite{Gao2023Retrieval}            & \XSolidBrush & \Checkmark &  \XSolidBrush & \XSolidBrush & \XSolidBrush &  \Checkmark & \XSolidBrush & \XSolidBrush & \Checkmark &  \Checkmark & \XSolidBrush & \XSolidBrush\\
& ~\cite{zhou2024trustworthiness}            & \XSolidBrush & \XSolidBrush &  \Checkmark & \Checkmark & \Checkmark &  \Checkmark & \Checkmark & \Checkmark & \Checkmark &  \Checkmark & \XSolidBrush & \XSolidBrush
\\ \cmidrule(l){2-14} 
& Ours                                                      & \Checkmark                                                               & \Checkmark                                                                          & \Checkmark                                                      & \Checkmark                                                                             & \Checkmark                                                 & \Checkmark                                                                  & \Checkmark                                         & \Checkmark                                                              & \Checkmark                                         & \Checkmark  & \Checkmark & \Checkmark                                        \\ \bottomrule
\end{tabular}
\end{table}

\subsubsection{Related Surveys and Differences}
As shown in Table \ref{survey-taxonomy}, recently, several surveys have been conducted on RAG~\cite{Gao2023Retrieval, fan2024survey} and trustworthy LLMs~\cite{trustworthy_llms_2024, huang2023surveysafetytrustworthinesslarge}. On one hand, although the RAG surveys provide a comprehensive overview of the state-of-the-art models and architectures, detailing some of the methods' efforts to address challenges in the realm of trustworthiness (e.g., robustness), few of them provide a comprehensive and focused discussion on the specific challenges and solutions related to trustworthiness across the entire RAG pipeline, particularly in areas such as reliability, fairness, privacy, and safety~\cite{Gao2023Retrieval, fan2024survey}. On the other hand, the surveys on LLM trustworthiness focus mostly on the generation aspects of LLMs, such as mitigating hallucinations or enhancing explainability, and cannot be directly applied to RAG models due to the added complexity introduced by the retrieval and augmentation stages~\cite{trustworthy_llms_2024}.

Most recently, there is a survey dedicated to trustworthiness in RAG systems~\cite{zhou2024trustworthiness}. Although the survey provides reviews on existing works, most of its content focuses on experiments related to trustworthy generation across different language models within a basic retrieval-augmented generation setup. We acknowledge the importance of such empirical studies in advancing the field but would like to emphasize the different focus of our survey. Rather than empirical studies, our survey centers on providing a more detailed and comprehensive literature review. We systematically categorize trustworthiness challenges and solutions across the RAG applications. By focusing on these broader aspects, our work aims to establish a unifying framework to guide future research and development in trustworthy RAG systems.

Nonetheless, all of the existing surveys have acknowledged the importance of trustworthiness and included it as a critical area for future research directions~\cite{Gao2023Retrieval, fan2024survey}, demonstrating the aligned interests in the community. 

\subsection{Paper Collection}
To construct a comprehensive survey of trustworthy RAG systems in LLMs, we follow a systematic
literature review methodology. We began by identifying relevant papers through
keyword searches across major academic databases, including Google Scholar, ACM
Digital Library, and arXiv, using general terms such as "Retrieval-Augmented
Generation," "trustworthiness," as well as specific terms for the different
aspects. Papers were included if they directly discussed the trustworthiness
aspects of RAG systems, such as reliability, privacy, safety, fairness, and accountability.
Because of the modular nature of RAG-based systems where external databases are
coupled with a language model, we also take works that discuss each into
consideration. After an initial screening, we categorize the papers into each
aspect of trustworthiness, situating them into corresponding taxonomies. Our
final collection of papers, as discussed in this survey, reflects a diverse
range of perspectives on RAG and trustworthiness. The cut off date for the papers to be included in this survey is October 2024. 

\subsection{Notes on the Organization of the Survey}
As outlined in the previous sections, the remainder of this survey is structured around the six key aspects of trustworthiness: Reliability, Privacy, Safety, Fairness, Explainability, and Accountability. Given the distinct nature of each aspect, every section will follow its own
taxonomy, introducing relevant works accordingly. To further emphasize these
differences, we include specific future directions and evaluation protocols
within each section. A general discussion of the future directions for
trustworthy Retrieval-Augmented Generation (RAG) systems will follow at the
conclusion, providing a broader discussion that encompasses and integrates the
section-specific insights.
\section{Reliability of Retrieval Augmented Generation}
While RAG improves factual consistency and adaptability, it also introduces unique reliability challenges. Unlike standalone generative models, RAG reliability depends not only on the underlying LLM but also the alignment of retrieved information. Ensuring reliability in RAG therefore requires evaluating both the the retrieval process and the generation conditioned on the retrieved content.

\subsection{Taxonomy of RAG Reliability}
At a high level, reliability requires the system to perform as expected under various conditions. Just as other deep learning models that train on the big data, RAG models are susceptible to common pitfalls of reliability. Previous work~\cite{tran2022plex} defines reliability from three granular aspects: the ability to express \textit{uncertainty} in predictions, the capability of \textit{robust generalization} under various conditions, and the extent to which the model can \textit{adapt} to new tasks. However, since RAG is inherently \textit{adaptable} because of the retrieved context, we will only consider \textit{uncertainty} and \textit{robust generalization} in our following discussion. 

\subsection{Uncertainty}
Uncertainty is a crucial factor for model reliability. Uncertainty quantification (UQ) helps quantify the
confidence in the model's predictions, which is essential in high-stakes
scenarios. Consider a medical question answering chatbot where a patient
inquires about their condition. If the model can express uncertainty in its
responses, it significantly reduces the risk associated with its predictions.
The patient can then make more informed judgments based on the confidence level
of the information provided. Thus, due to the imperativeness of accurately
conveying uncertainty, we need to ensure that robust uncertainty quantification
methods are integrated into the system. 

For Retrieval-Augmented Generation (RAG) systems, uncertainty quantification
presents two primary challenges: First, during the generation phase, uncertainty
stems from the inherent limitations of large language models (LLMs). Standard
techniques for quantifying uncertainty in LLMs, such as conformal prediction,
can be applied here with few adaptations~\cite{ye2024benchmarkingllmsuncertaintyquantification}. Second, uncertainty arises during the
retrieval phase and its interaction with the LLM, introducing a more complex
dynamics. The combination of retrieval and generation processes creates unique
challenges for UQ, necessitating advanced methods to
address the overall system complexity. The following section outlines ongoing
efforts to tackle these challenges, with a summary of the relevant literature
presented in Table \ref{uq-taxonomy}.

\begin{table}
\centering
\caption{Taxonomy for Uncertainty Quantification in RAG}
\label{uq-taxonomy}
\scriptsize
\begin{tabular}{ccccc}
\toprule
\textsc{Module} & \textsc{Reference}  & \textsc{White-box} & \textsc{Task} & \textsc{Year}\\
\midrule

\multirow{4}{*}{\textit{Generation}} 

 & Ye et al.~\cite{ye2024benchmarkingllmsuncertaintyquantification} & \XSolidBrush & Benchmarking & 2024 \\
 & Su et al.~\cite{su2024apienoughconformalprediction} & \Checkmark & Open Domain Question Answering & 2024 \\
& Kumar et al.~\cite{kumar2023conformalpredictionlargelanguage} & \Checkmark & Multiple Choice Question Answering & 2023 \\

& Quach et al.~\cite{quach2024conformallanguagemodeling} & \XSolidBrush & Open Domain Question Answering & 2023 \\

\midrule

\multirow{2}{*}{\shortstack{\textit{Retrieval} \\ + \textit{Generation}}} & Ni et al.~\cite{ni2024trustworthyknowledgegraphreasoning} & \XSolidBrush & Multi-hop Question Answering & 2024 \\
 & Li et al.~\cite{li2023traq}  & \XSolidBrush & Open Domain Question Answering & 2023\\

\bottomrule
\end{tabular}
\end{table}

\subsubsection{Uncertainty Quantification in Generation}
The generation phase in Retrieval-Augmented Generation (RAG) systems is critically influenced by the UQ of LLMs. Recent studies have explored various approaches in this area, with a focus on techniques like conformal prediction (CP) — a model-agnostic, distribution-free method that uses a calibration set to estimate prediction confidence~\cite{shafer2007tutorialconformalprediction}. To apply conformal prediction, a \textit{non-conformity score} is first defined to measure the confidence of a given prediction. Using a calibration set, the $1-\alpha$ quantile of the non-conformity score is then calculated, where $\alpha$ represents the user-defined error rate. Finally, the prediction set is constructed by selecting valid predictions based on the quantile score, ensuring that the set satisfies the $1-\alpha$ confidence level, assuming the calibration and test sets are exchangeable.

The cornerstone of CP lies in defining the \textit{non-conformity score}. In traditional multi-class classification, a common approach is to use the softmax score of the from the class prediction. Extending the logit-based non-conformity score to LLMs, methods have been further developed. Typically, they assume white-box access to the model, making them unsuitable for commercial LLMs such as ChatGPT. For instance, Kumar et al.~\cite{kumar2023conformalpredictionlargelanguage} applied standard CP to the Llama model~\cite{touvron2023llamaopenefficientfoundation} by leveraging softmax scores of token logits in multiple-choice tasks. Similarly, Ye et al.~\cite{ye2024benchmarkingllmsuncertaintyquantification} extended logit-based approaches to multiple baselines and language models.

To compensate the lack of application on black models, another promising direction is proposed for sampling-based techniques, where model confidence is estimated by repeatedly prompting the LLM. Quach et al.~\cite{quach2024conformallanguagemodeling} adapted the learn-then-test risk-control framework~\cite{angelopoulos2022learntestcalibratingpredictive} for LLMs, approximating the non-conformity score through sampling, which allows uncertainty quantification in black-box models without direct logit access. Su et al.~\cite{su2024apienoughconformalprediction} further advanced these methods by introducing non-conformity measures that integrate both coarse-grained and fine-grained notions of uncertainty, leading to smaller, more refined prediction sets.

\subsubsection{Uncertainty Quantification in Retrieval and Generation}
As shown in Figure \ref{fig:overview}, a traditional RAG system includes multiple components from retrieval to generation. Due to its complex, multi-component nature, directly applying LLM-based UQ methods
will produce less accurate, sub-optimal results~\cite{li2023traq, rouzrokh2024conflare}. This necessitates the
development of specialized techniques tailored to the unique structure and
requirements of RAG models. 

Recently, researchers proposed a multi-step calibration framework to enhance the retrieval process of RAG~\cite{rouzrokh2024conflare}. Specifically, this framework uses conformal prediction to quantify retrieval uncertainty, ensuring trustworthiness in RAG systems. The framework involves constructing a calibration set of questions answerable from the knowledge base and comparing their embeddings against document embeddings to identify the most relevant chunks containing the answers. By analyzing similarity scores and determining a cutoff threshold based on a user-specified error rate ($\alpha$), the system retrieves all chunks exceeding this threshold during inference. This multi-step calibration ensures the true answer is captured in the context with a (1 - $\alpha$) confidence level. 

Moreover, TRAQ~\cite{li2023traq} expanded the conformal prediction framework to include a Bayesian optimization module that minimizes the prediction set during the multi-step calibration. Because of the complexity of RAG, the constructed prediction set will be very large after aggregating the error rates of multiple components. By leveraging Bayesian optimization, the framework efficiently searches for the optimal parameters that reduce the size of the prediction set while maintaining the desired confidence level. TRAQ ensures that the retrieval process remains both accurate and computationally feasible, enhancing the overall reliability and performance of RAG systems.

Besides uncertainty quantification in the process of document retrieval, \textsc{UaG}~\cite{ni2024trustworthyknowledgegraphreasoning} attempted to address the gap in uncertainty quantification of knowledge graph reasoning. Unlike vector databases, knowledge graphs encode knowledge as triplets and include structural information. One representative task of knowledge graph reasoning is multi-hop question answering, where the system must infer answers by traversing multiple edges in the graph to connect the initial query node with the answer node. The \textsc{UaG} framework involves combining information from several related entities and relationships within the graph, further complicating the process of uncertainty quantification. \textsc{UaG} proposed to leverage a general risk control framework to find the optimal parameter for each stage of calibration, ensuring reliable uncertainty estimates while maintaining a reasonable prediction set size.

\subsection{Uncertainty Evaluation}

\paragraph{Metrics.} Traditionally, uncertainty quantification is evaluated from two key perspectives: \textit{coverage} and \textit{efficiency}~\cite{he2024surveyuncertaintyquantificationmethods}. Recall that the goal of uncertainty quantification is to ensure that the returned answer set satisfies a user-defined error tolerance of $1-\alpha$. Thus, the \textit{coverage rate} measures how effectively the model meets this requirement.

Given a returned set of answers, $\mathcal{A}_{\text{ret}}$, and the correct answer set, $\mathcal{A}_{\text{true}}$, the coverage rate, $C$, is calculated as the proportion of instances where the correct answer is included in the returned set. Formally, it is defined as:
\[
C = \frac{N_{\text{correct}}}{N_{\text{total}}},
\]
where $N_{\text{correct}}$ represents the number of times the correct answer $\mathcal{A}_{\text{true}}$ is contained in the returned set $\mathcal{A}_{\text{ret}}$, $\mathcal{A}_{\text{true}} \subseteq \mathcal{A}_{\text{ret}}$, and $N_{\text{total}}$ is the total number of instances. 

For the model to be considered reliable, $C$ should be at least $1-\alpha$. However, simply exceeding this threshold does not necessarily indicate optimal performance. Overestimating the returned set size while still satisfying the desired error rate implies inefficiency, as a smaller set could suffice for the same error rate.

Alongside coverage, \textit{efficiency}, denoted as $E$, is another critical metric, often evaluated by the size of the returned answer set (i.e. the number of returned answers per question):
\[
E = |\mathcal{A}_{\text{ret}}|.
\]
\noindent Efficiency reflects the utility of the model’s output, as larger sets may contain more irrelevant information, reducing their usefulness to the user. Thus, an efficient uncertainty quantification process minimizes $E$ while maintaining the desired coverage rate, $C$.

\subsection{Robust Generalization}
Previous work~\cite{tran2022plex} defines \textit{robustness} as the ability to make accurate estimates or forecasts about unseen events caused by out-of-distribution data, covariate shift, domain change, concept change, or population shift, etc. In the context of RAG, the most significant challenge is the shift in the distribution of the database. Realistically, the database will always be evolving, introducing new knowledge into the system. Without dedicated robustness measures, this can cause the model to underperform in various situations. Consequently, it is essential to develop approaches that allow the model to continually learn from new data and adjust its retrieval and generation processes accordingly such as in concept drifts.  Specifically, we will consider two aspects of robustness for RAG: resilience against irrelevant context and resilience against corrupted or misinformation contexts. It is worth mentioning that there is another type of context that we define as \textit{adversarially constructed corrupted context}. Sometimes they are closely related to \textit{corrupted context}, but due to their adversarial nature, we will consider them in Section \ref{sec:robustness} for Adversarial Robustness. This section will focus on the context that occurs \textit{organically} over time. 

\subsubsection{Irrelevant Context}

Fang et al.~\cite{adaptive_adversarial_rag_2024} considers the noise robustness of RAG with adaptive adversarial training. The paper explores three types of retrieval noises: (i) contexts that appear to be related to the query but do not contain the correct answer, (ii) contexts that are entirely unrelated to the query, and (iii) contexts that are thematically related to the query but include incorrect information. With the conclusion that type (i) and (iii) noise are the most misleading to the language models, the authors developed Retrieval-augmented Adaptive Adversarial Training (RAAT) to regulate the retrieval of noisy text. To improve the robustness under the noisy data, RAAT generates adversarial samples (noises) by considering the model's sensitivity to various types of noises and shows significant robustness improvement. The study further demonstrates that RAAT can be integrated seamlessly with existing RAG systems, enhancing their performance without substantial computational overhead. 

In addition, Yoran et al.~\cite{robust_rag_iclr_2024} further explores the negative impact of the retrieval of irrelevant context on the model performance. They argue that the negative impact of the irrlevant context is a result of the lack of training data with the retrieved passages. As a result, the brittleness to noisy passages is expected during inference. To address this observation, the author propose to finetune the language models on noisy contexts. Finetuning on this additional context allows the model to learn to differentiate between useful and distracting information and minimizing the negative effect of the irrelevant context. The experiment result on five different open domain datasets has shown significant improvements of robustness against irrelevant context for both single-hop and multi-hop retrieval based question answering. 

\subsubsection{Corrupted Context}
Recently, Xu et al.~\cite{dual_rag_2024} proposed a theoretical framework to explore the benefits and detriments of the RAG, in the situation where there's a discrepancy between the retrieved knowledge and the LLM knowledge. Specifically, they observed that the similarity between the RAG representation and the retrieved representation is bounded by the benefits and detriments, and the similarity is positively correlated with the value of benefits minus detriments. These results suggest that the similarities can be used as a proxy for the benefits and detriments of the RAG. Building upon the theoretical results, the author further proposed an interactive inference framework X-RAG that leverages the benefit of both worlds of retrieved knowledge and LLM knowledge.

\subsection{Robustness Evaluation}
\paragraph{Metrics.}
The evaluation of the model's robustness focuses on assessing its performance when noise is present in the data. Thus, the setup of the noisy data, which will be detailed in the dataset section, plays a key role in this evaluation. Exsting metrics outlined in \ref{sec:eval} will be applied to assess the model's performance. It is important to note that there are different reporting styles for these metrics in the context of model robustness. Some authors present standard tables comparing the proposed model's performance against baselines~\cite{adaptive_adversarial_rag_2024, dual_rag_2024}, while others report only the performance delta between the proposed fine-tuned model and the corresponding baseline for better visualization~\cite{robust_rag_iclr_2024}.

\paragraph{Datasets.} Currently, there is no widely-used benchmark for RAG robustness. To simulate real-world conditions and evaluate robustness, existing works create customized datasets that incorporate generated noise. Typically, a common QA benchmark (e.g., TriviaQA) 
is used, and during the retrieval process, noises are injected into the retrieved content. Depending on the problem setting (\textit{irrelevant context} vs. \textit{corrupted context}), the noise is either randomly selected or filtered using heuristic techniques~\cite{dual_rag_2024, robust_rag_iclr_2024}. These datasets attempt to replicate the type of challenges encountered in realistic environments where the retrieved information may not perfectly align with the query.

Recently, 
Fang et al.~\cite{adaptive_adversarial_rag_2024} proposed a benchmark for noise-robust RAG. For each QA instance, the proposed dataset includes three types of augmented retrieval noise: relevant retrieval noise, irrelevant retrieval noise, and counterfactual retrieval noise where the answer entity is intentionally incorrect, as well as the golden retrieval data. \textit{We recognize this as one of the first publicly available datasets for RAG robustness evaluation, and future works could benefit from using this for benchmarking.} 

\subsection{Future Directions of RAG Reliability}
Reliability remains an important challenge in the development of trustworthy RAG systems. While our current sections are structured independently for uncertainty and robustness, future research should aim for a more integrated approach that captures the intricate interactions between these aspects. These concepts are not necessarily exclusive; uncertainty quantification can help the model produce more trustworthy results when faced with less accurate or noisy contexts, while better robustness can reduce the model's overall uncertainty.

An integrated framework addressing both uncertainty and robustness could enhance the adaptability and robustness of RAG systems, particularly in complex real-world applications where the boundaries between irrelevant and corrupted contexts are blurred. By combining strategies from uncertainty quantification with robust robustness techniques, future research can explore models that are not only resilient to noise but also capable of dynamically adjusting their confidence levels based on the context quality.

Moreover, future efforts should extend current evaluation benchmarks to include more comprehensive benchmarks that reflect real-world conditions. Current benchmarks such as RAG-Bench~\cite{adaptive_adversarial_rag_2024} mark a good effort, but the data are manually filtered by rules, which limits the scope of the challenges they present. Real-world scenarios are more complex, with a greater variety of noise. By incorporating these complexities, new benchmarks could better simulate the unpredictable nature of real-world applications. With better benchmarks, researchers will be able to test models under various reliability scenarios where uncertainty and robustness challenges are intertwined. Expanding evaluation methodologies in this direction would push the frontier of research in reliable RAG systems and offer more actionable insights for practical deployments.

Finally, drawing from interdisciplinary fields like dynamic knowledge graphs and active learning could lead to the development of more adaptive RAG systems. In dynamically evolving environments, new knowledge introduced in external databases may create inconsistencies or conflicts with previously retrieved information, posing significant reliability challenges. These interactions would require RAG systems to not only generalize well but also quantify uncertainty effectively when knowledge shifts or drifts occur. Such advances would make RAG systems more robust and practical, improving their performance in dynamic, real-world scenarios.
\section{Privacy of Retrieval Augmented Generation}
\begin{table}
\centering
\caption{Taxonomy for RAG Privacy}
\label{privacy-taxonomy}
\scriptsize
\begin{tabular}{cccccc}
\toprule
 & \textsc{Reference}  & \textsc{Training} & \textsc{Tasks} & \textsc{Leakage} & \textsc{Year}\\
\midrule

\multirow{4}{*}{\textit{Attack}} & Zeng et al.~\cite{privacy_rag_2024} & \Checkmark & Document Extraction \& Training Data & Internal \& External & 2024 \\

 & Liu et al.~\cite{liu2024mask} & \Checkmark & Membership Inference Attack & External & 2024 \\
 
 & Cohen et al.~\cite{cohen2024unleashing} & \XSolidBrush & MIA \& Document Extraction & External & 2024 \\ 

 & Jiang et al.~\cite{jiang2024ragthief} & \XSolidBrush & Document Extraction & Internal \& External & 2024 \\
 
  & Peng et al.~\cite{peng2024data} & \Checkmark & Document Extraction & External & 2024 \\

\midrule

\multirow{2}{*}{\textit{Defense}} & Zeng et al.~\cite{zeng2024mitigating} & \Checkmark & External Database & External & 2024 \\
 & Zeng et al.~\cite{privacy_rag_2024}  & \Checkmark & Document Extraction & External & 2024\\

\bottomrule
\end{tabular}
\end{table}

Although privacy risks in LLMs have been extensively studied, RAG systems introduce additional complexities by leveraging external data. This integration poses new challenges in maintaining privacy, ensuring data integrity, and managing the overall trustworthiness of the RAG system. In this section, we will introduce the threat model concerning privacy leaks in RAG systems. We will then discuss the current efforts to address these challenges. Towards the end, we will explore potential future directions for enhancing the trustworthiness and privacy of RAG systems. 

\subsection{Taxonomy of RAG Privacy}
In Table \ref{privacy-taxonomy}, we outline the existing efforts in addressing the privacy issues present in the RAG systems. We will briefly introduce the relevant taxonomy in the following. 

\subsubsection{Training}
Training refers to whether the attack or defense requires \textit{prior} training on the data. Models that require training typically assume a distinct threat model compared to those that do not. When a model requires training, it often presumes white-box access to either the retriever or the language model, allowing attackers or defenders to fine-tune or adjust components of the RAG system to exploit or mitigate vulnerabilities. On the other hand, models that do not require training typically rely on prompt-based methods or zero-shot techniques. This distinction has significant implications for the feasibility of privacy attacks and defenses in RAG systems.

\subsubsection{Tasks} Admittedly, research on RAG privacy is still in its infancy. Current literature focuses on three main tasks: \textit{Document Extraction}, \textit{Training Data Extraction}, and \textit{Membership Inference Attack}. \textit{Document Extraction} seeks to extract confidential information from the retrieval database, such as Personally Identifiable Information (PII). \textit{Membership Inference Attack} aims to determine whether specific passages are present in the retrieval database. While not a direct privacy attack, it introduces risks by exposing sensitive associations between queries and database contents, potentially enabling adversaries to infer private information. Lastly, \textit{Training Data Extraction} examines the leakage of LLM training data in the context of retrieval-augmented generation, highlighting vulnerabilities that could lead to the unauthorized exposure of proprietary or sensitive datasets.

\subsubsection{Leakage}
We consider two sources of leakages. First, the leakage of the external retrieval database involves leaking targeted/untargeted information from external knowledge sources, such as sensitive data in proprietary databases or publicly available but privacy-relevant information inadvertently retrieved during query processing. Second, the leakage of the internal training data focuses on the exposure of the LLM training data in the context of retrieval-augmented generation. This occurs when the language model unintentionally reproduces sensitive or proprietary information from its training dataset during response generation, raising concerns about policy violations and privacy breaches. We organize the rest of the section from the above two aspects. 

\subsection{Data Leakage From the External Retrieval Database}

The goal of the attacker is to exploit privacy vulnerabilities within the retrieval dataset, targeting two main objectives: (1) eliciting specific information from the retrieval system with high accuracy, and (2) outputting the retrieved private data. Zeng et al.\cite{privacy_rag_2024} introduced a composite structured prompt, formulated as $q = {\text{information}} + {\text{command}}$, which leverages the context retriever's propensity for similarity-based matching. However, a significant limitation of this approach is its reliance on fixed queries, which cannot dynamically adapt to varying contexts. To address this limitation, Jiang et al.\cite{jiang2024ragthief} proposed a learning-based method. Their framework begins with an initial adversarial query and iteratively refines it based on the model's responses, progressively generating queries to extract as many documents as possible from the retrieval database.

When considering white-box access to the model, Peng et al.~\cite{peng2024data} focused on data extraction through backdoor attacks. Their method trains a model to associate specific triggers with desired outputs. Beyond directly extracting documents, their approach also explores generating stealthy outputs by employing a language model to paraphrase the retrieved content, thereby increasing the difficulty of detecting the attack. Furthermore, Cohen et al.~\cite{cohen2024unleashing} demonstrated that these attacks can escalate beyond isolated cases. By crafting an \textit{adversarial self-replicating prompt}, attackers can initiate a chain reaction that propagates through the entire Retrieval-Augmented Generation (RAG) system.

Although distinct from direct extraction methods and based on a different threat model, \textit{membership inference attacks} have also proven effective for data extraction. These attacks allow malicious users to infer whether specific content is present in the retrieval database. Liu et al.~\cite{liu2024mask} introduced a mask-based attack that obscures portions of documents, compelling the language model to predict the masked words. This technique not only reveals sensitive information but also highlights vulnerabilities in the retrieval system's training data.

\subsection{Data Leakage From the LLM Training Data}
The goal of the attacker is to extract data from the LLM's training and fine-tuning data that are encoded in the model parameters. In their paper, Zeng et al.~\cite{privacy_rag_2024} compared the effect of RAG in preventing data leakage from the LLM training data. The result shows that incorporating retrieved passages greatly reduces LLM's propensity to reproduce content memorized during its training/fine-tuning process. To isolate the effect of retrieval data integration, the author also attached 50 tokens of random noise injection as prefix. Although the random noise could also mitigate the data leakage, it is far less effective than integrating the retrieved content. 

\subsection{Defense on Privacy Attacks}
Although still a relatively under-explored area, some works have proposed defenses to mitigate privacy vulnerabilities in RAG systems. In their foundational work, Zeng et al.~\cite{privacy_rag_2024} observed that using a separate model to summarize the retrieved documents effectively reduces privacy leakage by abstracting sensitive information into generalized content. Additionally, they proposed implementing a distance threshold in the retrieval database, ensuring that only documents with certain relevance requirements are returned. However, this approach introduces a trade-off between system performance and privacy protection, as stricter thresholds can limit retrieval accuracy.

Building on these mitigation strategies, the authors further suggested the use of purely synthetic data as a way to entirely avoid potential leakage of real data~\cite{zeng2024mitigating}. This method involves identifying importing attributes of the data through few-shot samples, extracting key information associated with these attributes, and generating synthetic data that mirrors the original data without exposing sensitive information. This approach has shown promise in effectively mitigating privacy leakage while maintaining the performance of the RAG system.

\subsection{Privacy Evaluation}
\paragraph{Metrics.} Metrics for evaluating privacy attacks focus on quantifying the extent of information leakage and the effectiveness of extraction methods. Commonly used metrics include the total volume of retrieved context, the number of prompts that successfully yield substantial overlaps (e.g., at least 20 matching tokens) with the dataset, and the number of unique excerpts extracted. For targeted attacks, the evaluation centers on the precision of the extracted information, assessing how accurately specific targets are retrieved. In the case of untargeted attacks, metrics often rely on content similarity measures, such as ROUGE-L scores, to determine the degree of alignment between the retrieved content and the original dataset~\cite{privacy_rag_2024, zeng2024mitigating}.

\paragraph{Datasets.} Similar to the generalization evaluation, there haven't been any established datasets or baselines specifically designed for evaluating privacy in RAG systems. However, current work has leveraged existing datasets to provide initial insights. The Enron Email dataset~\cite{klimt2004enron} contains employee emails, which often include sensitive personal information, such as names, contact details, and internal company communications. The HealthcareMagic-101 dataset~\cite{zeng2024mitigating}, on the other hand, consists of doctor-patient dialogues, encompassing a variety of medical discussions that include personally identifiable information (PII) and private health information. These datasets serve as a valuable starting point for privacy evaluations due to their realistic, sensitive content that mimics the types of data RAG systems may encounter.

\subsection{Future Directions of RAG Privacy}
\paragraph{Domain Specific Applications.} While the preliminary findings from Zeng et al.~\cite{privacy_rag_2024} shed light on the dual nature of privacy risks in RAG systems, there remains a significant gap in understanding the broader implications of these vulnerabilities across different applications and contexts. Future research should prioritize developing robust, application-specific privacy-preserving techniques tailored to the unique demands of various domains, such as healthcare, finance, and legal services, where the consequences of privacy breaches can be particularly severe.

\paragraph{Privacy Defense.} Additionally, exploring the integration of advanced cryptographic methods, such as secure homomorphic encryption, within RAG frameworks could provide new avenues for safeguarding sensitive data during retrieval and generation processes. Another promising direction is the development of differential privacy techniques specifically adapted for RAG systems, aiming to balance privacy preservation with the utility of the generated outputs.

\paragraph{Evaluation.} Finally, as the field of RAG continues to evolve, it will be crucial to establish comprehensive benchmarks and evaluation metrics for privacy in these systems. These benchmarks should account for the diverse range of privacy threats, including both direct data leakage and more subtle inferential attacks, to ensure that the proposed solutions are rigorously tested and validated across a wide spectrum of scenarios. By addressing these challenges, future research can contribute to the creation of more secure and trustworthy RAG systems, ultimately fostering greater confidence in the deployment of these technologies in sensitive and high-stakes environments.

\section{Safety of Retrieval Augmented Generation}  
\label{sec:robustness}

Recent research has shown that LLMs are vulnerable to a range of adversarial attacks~\cite{huang2023surveysafetytrustworthinesslarge, shayegani2023surveyvulnerabilitieslargelanguage, liu2024towards, Yao_2024}. Through techniques such as prompt engineering, hint manipulation, and input perturbation, attackers can bypass safety mechanisms and exploit model weaknesses, posing significant threats to society. RAG systems, built upon the capabilities of LLMs and integrated with external databases, present distinct challenges related to safety concerns. For example, by letting the RAG system retrieve adversarial information, attackers can circumvent the alignment of LLMs with human integrity and produce malicious content~\cite{deng2024pandorajailbreakgptsretrieval}. As more adoption of RAG system emerges, the safety of RAG systems have arisen significant concerns on utilizing RAG in more high-stake applications. For example, in education applications, RAG is often leveraged to retrieve relevant domain-specific educational context (textbooks, problem set, etc.). Thus, these vulnerabilities will pose significant threats to underage minors if the system is adversarially compromised. As a result, developing robust RAG is essential for ensuring trustworthiness in RAG systems. Since this is a relatively new field of research, we will give a brief overview of the adversarial attacks on RAG and then point out potential future directions that are worth investigating. 

\subsection{Taxonomy of RAG Safety} In Table \ref{robustness-taxonomy}, we summarize existing Retrieval-Augmented Generation (RAG) methods based on an adversarial taxonomy. This section introduces the taxonomy and provides definitions for each category. It is worth noting that the wide range of adversarial attacks on LLMs, such as backdoor attacks~\cite{xue2024trojllm, lu2024test}, jailbreaking attacks~\cite{wei2024jailbroken, zou2023universal}, and prompt injection attacks~\cite{greshake2023not, liu2023prompt, yan2023backdooring}, technically target the underlying LLM component of RAG. However, how the retrieved context influences the attack surface and defense strategies in RAG systems remains an open question requiring further investigation. Preliminary results~\cite{privacy_rag_2024} suggest that RAG can alleviate the effects of simple prefix attacks, as the retrieval step introduces an additional layer of complexity. Yet, the interaction between the retriever and generator 
under more complex adversarial conditions, e.g., combined backdoor and retrieval-based attacks, is still poorly understood warranting further study.

\begin{table}
\centering
\caption{Taxonomy for RAG Safety}
\label{robustness-taxonomy}
\scriptsize
\begin{tabular}{ccccc}
\toprule
 & \textsc{Reference}  & \textsc{White-Box} & \textsc{Black-Box}  & \textsc{Year}\\
\midrule

\multirow{4}{*}{\textit{Targeted}} & Zou et al.~\cite{zou2024poisonedragknowledgecorruptionattacks} & \Checkmark & \Checkmark  & 2024 \\

 & Xue et al.~\cite{xue2024badrag} & \Checkmark & \XSolidBrush & 2024 \\
 
 & Long et al.~\cite{long2024backdoorattacksdensepassage} & \Checkmark & \XSolidBrush & 2024 \\ 

 & Zhong et al.~\cite{zhong2023poisoningretrievalcorporainjecting} & \Checkmark & \XSolidBrush & 2023 
 \\
 \midrule
\multirow{2}{*}{\textit{Jailbreak}} & Wang et al.~\cite{wang2024poisonedlangchainjailbreakllms} & \XSolidBrush & \Checkmark & 2024 \\
 & Deng et al.~\cite{deng2024pandorajailbreakgptsretrieval}  & \XSolidBrush & \Checkmark & 2024\\

\bottomrule
\end{tabular}
\end{table}

Due to the scope of this survey, we do not delve into existing LLM-specific attacks, which are covered extensively in works like~\cite{trustworthy_llms_2024, huang2023surveysafetytrustworthinesslarge}. Instead, our focus in this section is on the robustness of the retriever model in RAG systems and how its interaction with the generator model can impact overall system security and resilience to adversarial attacks.

\subsubsection{Threat Model} To systematically categorize the robustness of RAG systems, we consider three primary components: the external database, the output generator (the underlying language model), and the context retriever. A realistic threat model assumes that the attacker has no read or delete access to the external database but possesses write access. This mirrors real-world scenarios where users can upload documents to a database but cannot access all of the content. The attacker is also assumed to have no detailed knowledge of the underlying language model. In practice, most commercial language models are proprietary, making them black-box systems. This assumption is crucial for adversarial robustness studies, as the attacker cannot exploit specific weaknesses in the model architecture or training data. 

For the retriever, we consider two distinct threat models. For white-box setting, the attacker has complete access to the retriever model. This includes the ability to inspect the model architecture, parameters, and sometimes retriever-specific training data. The attacker can thus craft sophisticated adversarial examples by exploiting the retriever’s known weaknesses. For example, by understanding the tokenization or ranking algorithm, an attacker might introduce documents that are highly ranked by the retriever but irrelevant or misleading for the generation task~\cite{zou2024poisonedragknowledgecorruptionattacks}.

In contrast, the black-box setting assumes the attacker has no direct access to the retriever. The adversary can only query the retriever and observe the output (i.e., the ranked retrieved documents) without knowledge of the internal mechanisms. The attacker must infer patterns from these outputs and attempt to manipulate the retriever's behavior through indirect methods, such as poisoning the external database or introducing misleading or noisy entries~\cite{deng2024pandorajailbreakgptsretrieval}.

\subsubsection{Attacker's Goal} The attacker's target, under traditional machine
learning contexts, is often categorized into two main types: targeted attacks
and untargeted attacks~\cite{trustworthy_graph_neural_networks_2024,
trustworthy_llms_2024}. For generative models such as Retrieval-Augmented
Generation (RAG), we identify two corresponding categories: targeted attacks and
jailbreak attacks. 
Targeted attacks are aimed at
manipulating the model's output in response to specific inputs, usually focusing
on certain questions or topics. The goal is to subtly distort the generated
response while keeping the attack as stealthy as possible, making it harder to
detect through typical monitoring systems. These methods are often carefully
crafted to evade detection mechanisms by introducing minimal disruptions. Such
targeted attacks can have significant social implications. For example, by
injecting retrieval bias, a targeted attack might skew the model's output to
emphasize certain policies or downplay others, inadvertently affecting sensitive
areas such as elections. 

In the context of generative models, untargeted attacks manifest as jailbreaking attacks, where attackers attempt to bypass content restrictions or safety measures embedded within the model. The goal is to provoke unrestricted or harmful outputs without focusing on a specific topic. This type of attack poses a broad threat as it can force the model to generate inappropriate or unsafe content across a range of inputs, compromising the system’s reliability and trustworthiness.

\subsection{Methods of RAG Safety}
\paragraph{Targeted Attacks.} 
Zou et al.~\cite{zou2024poisonedragknowledgecorruptionattacks} first introduce PoisonedRAG as a novel method to attach the new attack surface brought by the retrieval component. By designing specific passages to be injected into the retrieval database based on specific questions, PoisedRAG is able to mislead the RAG system to generate specific answers desired by the attackers for specific questions. It considers both the white-box and black-box settings. When the attacker has no access to the model parameters (black-box), PoisonedRAG crafts the injected passage with a simple heuristic: passages that is more similar to the question would be more likely to be retrieved. On the other hand, when the attacker has access to the model parameters, the crafted message is constructed by further optimizing the following equation 
\begin{equation}
    P = \underset{P'}{\text{argmax}}Sim(f(Q), f(P'))
\end{equation}
where $P$ is the generated passage, $Q$ is the user query, $f$ is the encoder, and $Sim$ is the function that measures the similarity between the encoded passage and question.

One limitation of PoisonedRAG is its focus on specific queries, neglecting broader group-based attacks that target semantically related query categories, such as those involving political affiliations, race, or religion. To address this gap, Xue et al.~\cite{xue2024badrag} propose the BadRAG framework, which extends the attack methodology to include group-query targeting. BadRAG allows the trigger to be semantic groups such as political parties or candidates by collecting target triggers from the given topic. For example, for \textit{Republican}, BadRAG collects terms including \textit{Governor}, \textit{Red States}, and \textit{Pro-Life}. To optimize the adversarial passages, BadRAG employs a contrastive learning paradigm, where the triggered queries are positive samples and the normal queries are negative samples. The adversarial passage will then maximize its similarity with the triggered queries and minimize its similarity with the normal ones. It's also worth mentioning that this approach can also support additional attacks, such as Denial of Service (DoS)~\cite{xue2024badrag}, by aligning adversarial passages with targeted model behaviors.

\textit{Dense retrieval} has been extensively studied within the Information Retrieval community~\cite{densesurvey2024}. Many attacks on dense retrievers, such as adversarial manipulation, are also applicable to passage retrieval tasks. Recently, Long et al.\cite{long2024backdoorattacksdensepassage} introduced a backdoor attack framework that exploits grammatical errors as triggers to spread misinformation. By employing contrastive learning to fine-tune the retriever, the model can retrieve adversarial passages specified by an attacker when it detects these grammatical anomalies. Additionally, Zhong et al.\cite{zhong2023poisoningretrievalcorporainjecting} demonstrated that adversarial passages trained on one domain can effectively transfer to out-of-domain queries, broadening the scope and potential impact of such attacks.

Despite these advances, most attack strategies for dense retrieval have been developed without considering downstream generation tasks, leaving their effects on generated outputs unclear. For instance, certain Retrieval-Augmented Generation (RAG) methods equipped with safety guardrails could potentially diminish the impact of adversarially retrieved passages. As we will discuss in the Future Directions section, understanding and addressing the interaction between retrieval attacks and downstream generation presents a significant research opportunity in Trustworthy RAG. 

\paragraph{Jailbreak Attacks.} When specific attack targets are absent, the threat model shifts toward jailbreak attacks. Wang et al.\cite{wang2024poisonedlangchainjailbreakllms} examine jailbreaking in the context of LangChain, a popular RAG framework. They analyzed jailbreak vulnerabilities in major Chinese Large Language Models and introduced the Poisoned-LangChain (PLC) method. By embedding jailbreak prompts into the retrieval database, PLC achieved jailbreak success across three scenarios, maintaining a consistent success rate exceeding 80\%. More recently, Deng et al.\cite{deng2024pandorajailbreakgptsretrieval} introduce Pandora, which extends jailbreaking attacks to English-based LLMs and more generalized RAG frameworks. Pandora enhances the malicious prompts by categorizing them into distinct topics and storing them in PDF format. This approach ensures that only titles and abstracts are retrieved, by which circumventing potential defense mechanisms that might detect the malicious content.

\subsection{Safety Evaluation}
\paragraph{Metrics.} For targeted attacks, researchers typically evaluate results from two perspectives. First, they measure the exclusivity of the trigger query’s effectiveness. To avoid detection, it’s essential that the same adversarial effects do not occur for non-triggered queries. To quantify this, retriever-based methods like BadRAG~\cite{xue2024badrag} assess the proportion of adversarial passage retrievals for clean queries compared to triggered queries. Specifically, they report the percentage of queries that retrieve at least one adversarial passage in the top-$k$ results (where $k=1, 10, 50$). Second, they measure the effectiveness using the Attack Success Rate (ASR). Notably, in generative tasks, ASR requires nuance due to variations in language expression. For example, responses like \textit{“Sam Altman”} and \textit{“The CEO of OpenAI is Sam Altman”} both correctly answer \textit{“Who is the CEO of OpenAI?”} Thus, researchers often employ \textit{substring matching} rather than \textit{exact matching} for this evaluation~\cite{zou2024poisonedragknowledgecorruptionattacks}.

Jailbreak attacks are also evaluated using ASR. However, lacking a targeted question, the criteria for successful jailbreaks need to be carefully defined. Deng et al.\cite{deng2024pandorajailbreakgptsretrieval} manually label a generation as a successful attack based on the \textit{relevance} and \textit{quality} of the generated content. Similarly, Yang et al.\cite{wang2024poisonedlangchainjailbreakllms} also manually count successful attacks to calculate ASR. We identify this as a methodological gap and will further discuss it in the future directions section.

\paragraph{Datasets.}
Currently, there are no widely accepted standardized datasets for evaluating robustness in retrieval-augmented generation (RAG) systems. For targeted attacks, researchers often follow a structured paradigm: first, identifying the downstream task. The current evaluations primarily focus on question answering, leveraging widely used benchmark datasets such as Natural Questions (NQ)\cite{kwiatkowski-etal-2019-natural}, MS MARCO\cite{bajaj2018msmarcohumangenerated}, and SQuAD~\cite{rajpurkar2016squad}. Researchers then select questions based on the specific characteristics of the targeted attacks. For instance, in BadRAG, Xue et al.~\cite{xue2024badrag} chose \textit{Republican} and \textit{Democrats} as targets for sentiment steering. Evaluation involves comparing results between clean (untargeted) queries and targeted queries to measure the impact of the attack.

For jailbreak attacks, evaluation methods vary significantly due to the inherent challenges of assessing open-ended text generation. Current approaches typically rely on manually curated adversarial questions categorized into specific themes. For example, Pandora~\cite{deng2024pandorajailbreakgptsretrieval} organizes questions into categories such as \textit{Adult}, \textit{Harmful}, \textit{Privacy}, and \textit{Illegal}, whereas Poisoned-Langchain~\cite{wang2024poisonedlangchainjailbreakllms} uses categories like \textit{Dangerous Behaviors}, \textit{Misuse of Chemicals}, and \textit{Illegal Discrimination}. Despite these efforts, the RAG literature lacks standardized datasets and unified evaluation frameworks, highlighting the need for future research to establish comprehensive benchmarks and methodologies.

\subsection{Future Directions of RAG Safety}
\label{subsec:robust_future}
\paragraph{Adversarial Defense.}
Current defense mechanisms against adversarial attacks in retrieval-augmented generation (RAG) systems are rudimentary. For instance, Xue et al.~\cite{xue2024badrag} propose learning the connection between trigger words and adversarial passages through \textit{token masking}. However, there is a notable lack of dedicated research specifically addressing adversarial defense in this domain. Existing approaches are limited in their scope and sophistication, leaving significant room for improvement. Arguably, advancements in generalization research, as discussed in Section 3.2, can be leveraged for adversarial defense. Training RAG models on distributions that include retrieved adversarial data could enable these models to better generalize and potentially mitigate the impact of adversarial passages. However, further exploration is needed to design and implement targeted approaches that address the unique challenges posed by adversarial attacks in RAG systems. The development of dedicated methodologies for adversarial defense remains an essential and largely uncharted area of research.

\paragraph{Modality.}
Current research predominantly focuses on RAG systems operating over text databases represented as vectors. However, recent advancements have introduced emerging applications that utilize alternative knowledge representations, such as knowledge graphs and databases~\cite{ni2024trustworthyknowledgegraphreasoning}. The unique structural properties of knowledge graphs present novel opportunities and challenges for both adversarial attacks and defenses. Attack and defense mechanisms tailored to these modalities are necessary to account for their inherent characteristics, such as the interconnectedness of entities and the semantic richness of relationships. Future research should focus on developing methods that address these unique requirements to expand the applicability and robustness of RAG systems across diverse knowledge representations.

\paragraph{Evaluation.}
As highlighted in preceding sections, the field currently suffers from a lack of standardized evaluation protocols. This absence hinders the ability to conduct fair comparisons and benchmark the effectiveness of different approaches. We advocate for the establishment of comprehensive evaluation frameworks and benchmarks that consider diverse metrics, such as robustness, generalization, and performance under adversarial conditions. Such benchmarks would provide a unified basis for assessing advancements in the field and drive progress through consistent and meaningful comparisons. Addressing this gap is critical for fostering innovation and ensuring the reliability of RAG systems in practical applications.

\section{Fairness of Retrieval Augmented Generation}
As generative models such as LLMs and image-generative models becomes increasingly integrated into real-world applications, it is critical to ensure fairness in their outputs. Retrieval-Augmented Generation (RAG) systems combine generative models with external knowledge retrieval, providing substantial improvements in accuracy and relevance by incorporating up-to-date information. However, the external data retrieved by these models may contain societal biases due to biased pre-trained knowledge \cite{wang2022revise, birhane2021large, birhane2021multimodal}, which leads to biased outputs. This further introduces the risk of amplifying disparities in age, gender, race, and other demographic attributes, particularly when the retrieved data is drawn from biased or unregulated sources.

\subsection{Taxonomy of RAG Fairness}
 In this section, we investigate various approaches that aimed at promoting fairness in RAG models. As shown in Table \ref{fairness-taxonomy}, Ensuring fairness in Retrieval-Augmented Generation (RAG) systems requires addressing biases at two stages: the retrieval of external data and the generation of outputs. These stages introduce unique fairness challenges, from biased data sources to unfair generative behavior, which need to be mitigated to create equitable and unbiased systems.

\subsection{Fairness in Retrieval}
One major challenge for fairness in RAG systems lies in the retrieval phase, where external knowledge or data used to enhance generation is sourced. Fairness issues during this stage can arise from various factors, including the retrieval model, the retrieval process, and the re-ranking mechanism. 
To address these challenges, several frameworks have been proposed to ensure that the retrieval system itself is fair.
Reskabsaz et al.~\cite{rekabsaz2020neural} introduces a framework for measuring bias that quantifies gender-related biases in ranking lists and assesses the impact of both BM25 and neural retrieval models. Furthermore, Reskabsaz et al.~\cite{rekabsaz2021societal} investigates how re-ranking methods can mitigate biases present in initial retrieval results. Then, Wang et al.~\cite{wang2024large} recognizes a gap between fairness and ranking performance when using LLMs for re-ranking and proposes a method with LoRA. To ensure demographic diversity, FairRAG ~\cite{shrestha2024fairrag} incorporates external data sources that cover a broad range of age, gender, and skin tone categories. This approach uses post-hoc sampling techniques to debias the retrieval process, preventing disproportionate representation of specific demographic groups in the retrieved data.
Beyond addressing bias in the data itself, frameworks such as BadRAG~\cite{xue2024badrag} have shown how maliciously inserted or poisoned data in the retrieval corpus can lead to biased and unfair outputs. Kong et al.~\cite{kong2024mitigating} proposes the Post-hoc Bias Mitigation (PBM) technique, which balances retrieved image sets to ensure more equitable representation across gender and race.

\begin{table}
\centering
\caption{Taxonomy for RAG Fairness}
\label{fairness-taxonomy}
\scriptsize
\begin{tabular}{ccccc}
\toprule
\textsc{Module} & \textsc{Reference}  & \textsc{Bias Mitigation} & \textsc{Focus} & \textsc{Year}\\
\midrule

\multirow{4}{*}{\textit{Retrieval}} 

& Wang et al.~\cite{wang2024large} & LoRA Fine-tuning & Ranking Fairness vs. Performance & 2024 \\

& Shrestha et al.~\cite{shrestha2024fairrag} & Diverse Sampling & Demographic Diversity & 2024 \\

& Rekabsaz et al.~\cite{rekabsaz2021societal} & Re-ranking Methods & Societal Bias Mitigation & 2021 \\

& Rekabsaz et al.~\cite{rekabsaz2020neural} & Post-hoc Re-ranking & Gender Bias in Retrieval & 2020 \\

\midrule

\multirow{4}{*}{\textit{Generation}} 

& Wu et al.~\cite{wu2024does} & Empirical Evaluation & Cross-task Fairness & 2024 \\

& Wang et al.~\cite{wang2023decodingtrust} & Output Conditioning & GPT-3.5/4 Bias Detection & 2023 \\

& Liang et al.~\cite{liang2022holistic} & Representation Adjustment & Fair Question Answering & 2022 \\

& Parrish et al.~\cite{parrish2021bbq} & Benchmark Evaluation & Stereotype Analysis & 2021 \\

\midrule

\multirow{3}{*}{\shortstack{\textit{Retrieval} \\ + \textit{Generation}}} 
& Kong et al.~\cite{kong2024mitigating} & Post-hoc Bias Mitigation (PBM) & Gender and Race Fairness & 2024 \\

& Kim et al.~\cite{kim2024towards} & Fair Retrieval + Generation & Fairness-Quality Trade-off & 2024 \\

& Shrestha et al.~\cite{shrestha2024fairrag} & Cross-Modal Guidance & Demographic Balancing & 2024 \\

\bottomrule
\end{tabular}
\end{table}

\subsection{Fairness in Generation}

Once the data is retrieved, the next challenge lies in ensuring that the generative process itself is fair. Even with fair retrieval, generative models may introduce biases based on how the retrieved data is integrated. To promote fairness in generation, Liang et al.~\cite{liang2022holistic} assesses the accuracy of question-answering systems while accounting for fairness through measures of toxicity and representation bias. Similarly, Wang et al.~\cite{wang2023decodingtrust} identifies the demographic imbalances in models like GPT-3.5 and GPT-4 under both zero-shot and few-shot question-answering settings. FairRAG~\cite{shrestha2024fairrag} employs conditioning techniques where generative models are guided by references that are demographically diverse. By incorporating external images or data from a wide range of demographic groups, these models produce more balanced and representative outputs. Parrish et al.~\cite{parrish2021bbq} introduces the BBQ benchmark to evaluate biases in LLM-generated responses by examining the reliance on stereotypes and anti-stereotypes in both ambiguous and disambiguated contexts. Next, to fully explore fairness throughout all stages and components of RAG pipelines, Wu et al.~\cite{wu2024does} conducts an empirically evaluation of fairness across various RAG methods. Similarly, Kim et al.~\cite{kim2024towards} evaluates RAG systems with a fairness-aware retriever across seven different tasks and identifies the overall trend of fairness-quality trade-off, considering both retrieval and generation performance.

\subsection{Fairness Evaluation}
\paragraph{Metrics.} First, to assess the accuracy of generated answers, it is common to use Exact Match (EM)~\cite{rajpurkar2016squad} and ROUGE-1 scores~\cite{lin2004rouge}. For fairness evaluation, the focus is on metrics such as Group Disparity (GD)~\cite{friedler2019comparative} and Equalized Odds (EO)~\cite{hardt2016equality}. Group Disparity measures the performance difference between protected and non-protected groups by calculating the ratio of exact matches within each group to the total number of exact matches across all groups. In contrast, EO evaluates whether the likelihood of correct answers (true positives) and incorrect answers (false negatives) is similar across different demographic groups, ensuring that no group is disproportionately advantaged or disadvantaged. Works like BadRAG~\cite{xue2024badrag} identify vulnerabilities and attacks on retrieval components (RAG database) and their indirect effects on generative parts (LLMs). They evaluate metrics such as retrieval success rate and rejection rate to assess the system's robustness and ensure that biases or attacks in retrieval do not impact generative outputs.

\paragraph{Datasets.} A popular dataset is the TREC Fair Ranking Track~\cite{ekstrand2023overview, craswell2020overview}, which includes subsets such as gender and location. The track aims to provide a platform for participants to develop and evaluate novel retrieval algorithms that ensure fair exposure to a mix of demographics or attributes, such as ethnicity, represented by relevant documents in response to a search query. The BBQ dataset~\cite{parrish2021bbq} includes samples with contexts that are either ambiguous or unambiguous. Ambiguous contexts test model behavior with insufficient evidence by providing only a general setting, while disambiguated contexts offer enough details to identify the correct individual for negative/non-negative questions. The LaMP benchmarks~\cite{salemi2023lamp} include various prediction tasks like classification, regression, and generation, and are ideal for scenarios where multiple items can be relevant, unlike typical QA tasks. With clear item providers and consumers, LaMP aligns with the goal of ensuring fairness for item providers and evaluates language models' personalization capability through retrieval-augmentation of user interaction histories.

\subsection{Future Directions of RAG Fairness}

\paragraph{Personalized Fairness.} While existing frameworks like FairRAG~\cite{shrestha2024fairrag} and PBM~\cite{kong2024mitigating} have made notable strides in mitigating biases in Retrieval-Augmented Generation (RAG) systems, they largely overlook the need for personalized fairness mechanisms. Personalized fairness involves tailoring fairness constraints to specific application contexts, such as healthcare or recruitment, where fairness definitions and requirements can vary significantly. For example, in healthcare, equity might require prioritizing underrepresented groups in clinical trials, while in recruitment, it might focus on reducing gender or racial biases in candidate evaluation. Future research should explore adaptive fairness constraints that dynamically align with these domain-specific requirements, ensuring equitable outcomes in high-stakes domains.

\paragraph{Fairness-Accuracy-Relevance Trade-offs.} A persistent challenge in advancing fairness in RAG systems is the inherent trade-offs between fairness, relevance, and accuracy. Current approaches, such as those explored by Liang et al.~\cite{liang2022holistic} and Wang et al.~\cite{wang2023decodingtrust}, primarily emphasize fairness in the generation and retrieval phases. However, they lack robust mechanisms to balance these competing objectives. Multi-objective optimization techniques could be pivotal in navigating these trade-offs. By formalizing fairness, relevance, and accuracy as interconnected yet distinct objectives, researchers can develop frameworks that prioritize trade-off management, enabling more holistic and practical solutions for fairness in RAG systems.

\paragraph{Cross-Modal Fairness.} The emergence of multimodal RAG systems that process both text and image data introduces unique challenges in fairness. Biases in these systems can manifest differently across modalities, yet current research, such as Wu et al.~\cite{wu2024does} and Kim et al.~\cite{kim2024towards}, has only scratched the surface of cross-modal fairness. Future work should aim to identify and mitigate modality-specific biases while ensuring consistent fairness across both text and image outputs. For example, ensuring that a system’s fairness constraints for text descriptions align with fairness principles for corresponding visual elements could significantly enhance trustworthiness. Addressing cross-modal fairness is essential for building robust and equitable multimodal RAG systems.
\section{Explainabilty of Retrieval Augmented Generation}
Explainability, as a critical perspective of trustworthiness, has attracted significant attention due to its capability to elucidate the behavior of machine learning models and uncover novel data-driven insights that can even inspire domain experts~\cite{selvaraju2020grad, robnik2018perturbation, ribeiro2016should, liu2023fair, feng2023degree, ying2019gnnexplainer}. This demand for explainability has been further intensified with the rise of black-box LLMs~\cite{liang2022holistic, sudhi2024rag}. We acknowledge the existing survey~\cite{zhou2024trustworthiness} on trustworthiness, which includes a section aimed at uncovering the decision-making process of RAG systems called transparency. However, the concept of transparency in that context differs from the explainability we focus on in our work. While transparency is more general and seeks to understand the algorithms and underlying rationales, our explainability (of the output) specifically aims to elucidate why a particular input (transparency) leads to a given output through a specific model (interpretability).

In the context of RAG, explainability becomes even more crucial due to the inherent complexity of its multi-component architecture. Beyond explaining the generation process of LLMs, it is also important to understand why the retrieval process prefers specific contexts over others. For example, what words in the input questions lead to the retrieval of a particular sentence? As some of the RAG systems include post-retrieval~\cite{glass2022re2g,kim2024sure,yang2023prca}, it is also important to understand why the retrieved content needs to be post-processed in a specific way to augment the downstream generation. For example, what features of the input result in such ranking/importance scores when performing reranking after retrieval? Given the inherently multi-stage nature of RAG, this section reviews the literature on explaining the two crucial stages of RAG, retrieval and generation process, respectively.

\begin{table}
\centering
\caption{Taxonomy for RAG Explainability}
\label{exp-taxonomy}
\scriptsize
\begin{tabular}{cccc}
\toprule
\textsc{Module} & \textsc{Reference}  & \textsc{Task} & \textsc{Year}\\
\midrule

\multirow{3}{*}{\textit{Generation}} & Sudhi et al.~\cite{sudhi2024rag} & English and German QA & 2024 \\

& Luo et al.~\cite{luo2024rog} & Knowledge Graph QA & 2024 \\

 & Rorseth et al.~\cite{rorseth2024rage} & Open-book QA & 2024 \\

\midrule

\multirow{3}{*}{\shortstack{\textit{Retrieval} \\ + \textit{Generation}}}
 &  Kunze et al.~\cite{tekkesinoglu2024feature}  & Scene-Understanding & 2024
 \\
 &  Hussien et al.~\cite{hussien2024rag}  & Road User Intention Explanation & 2024
 \\
 & Ferraretto et al.~\cite{ferraretto2023exaranker} & Document Retrieval & 2023 \\
\bottomrule
\vspace{1ex}
\end{tabular}
\end{table}

\subsection{Taxonomy of RAG Explainability}

\subsubsection{Explainability in Retrieval}
As illustrated in Table \ref{exp-taxonomy}, we situate our discussion of explainability in RAG in the distinction between retrieval, generation, and dual enhancement. To the best of our knowledge, no dedicated research efforts have been made to explain retrieval within the context of RAG. However, several studies have indeed explored explainability in the general information retrieval, particularly in recommender systems and search~\cite{zhang2020explainable,zhang201919,zhang2018sigir}. Therefore, we provide a high-level summarization of representative explanation techniques in information retrieval, with the expectation of inspiring similar success in explaining the retriever of RAG.

Based on~\cite{anand2023explainable}, the explanation methods in information retrieval can be categorized into post-hoc explanation, axiomatic strategies, probing strategies, and self-interpretable designs. The post-hoc explainers explain the models after they make decisions, the representative examples of which used in information retrieval are feature attribution and generative approach. The feature attribution works by ascribing the retrieved outcomes to certain input (i.e., the attribution). Some of the methods find the explanation features by computing the feature importance, such as~\cite{qiao2019understanding} uses interpretable textual features to explain rankings,~\cite{polley2022towards} understands the BERT-based ranking models by the attention scores of tokens, and~\cite{verma2019lirme} estimates the point-wise explanations by analyzing the contribution of each token to the output of the ranking model. Other methods try to explain the model outputs by finding the most explanatory features, e.g.~\cite{singh2021extracting} uses a greedy search-based algorithm to obtain a subset of features that serve as the explanations. Axiomatic explainers provide explanations using axioms~\cite{volske2021towards} and probing explainers provide valuable insights into the inner workings of neural models by revealing what types of information are encoded in their embeddings and model parameters~\cite{cohen2018understanding}, how sensitive they are to various textual properties~\cite{macavaney2022abnirml}, and what knowledge they possess~\cite{choi2022finding,formal2021white}. Although self-interpretable explainers inherently provide explanations by their designs, making the model fully transparent is extremely challenging, and usually, only specific components are interpretable and transparent~\cite{formal2021white,zhang2021explain}. 

\subsubsection{Explainability in Generation}
In addition to explaining the retriever, explaining the generator is equivalently or even more important. First, the output directly stems from this generation process, and any shortcuts or reliance on misleading features can lead to significant generation errors~\cite{hong2023so, deng2024cram}. Second, modern RAG systems typically rely on LLM-based generators, which are prone to hallucinations. This raises concerns about whether the generated output truly focuses on the query-relevant words and the retrieved contents, which can be naturally addressed via explanation~\cite{schneider2024explainable, sudhi2024rag}. There are two main categories of explanation methods in the generation process of RAG systems, ante-hoc explanation methods and post-hoc explanation methods, respectively. 

RAG-Ex~\cite{sudhi2024rag}, as a post-hoc explanation method, introduces a model and language-agnostic framework inspired by the philosophy of perturbation-based explanation. The core idea is to identify critical tokens or features such that removing them would significantly alter the output of the generator. Specifically, RAG-Ex proposes six perturbation methods: leave-one-token-out, random noise, entity manipulation, antonym injection, synonym injection, and order manipulation. After prompting LLMs with perturbed inputs (using the same parameters), the similarity between the generated responses and the original response is measured, which can be further used to calculate the importance scores for the perturbed tokens. After obtaining the importance score for each token, top-K important tokens are selected as the explanation, and they are referred to with the sentence containing the ground-truth answer or question to further assess the explanation quality. Through experiments, RAG-Ex finds that the "leave-one-token-out" perturbation performs the best in terms of accurately identifying the most critical tokens for the explanation. Aiming to trace the origin of LLM answers within the context of RAG, ~\citet{rorseth2024rage} proposed a post-hoc explainer called RAGE, which deduces the provenance and salience of external knowledge used during RAG. This framework is designed to generate counterfactual explanations for LLM answers by employing different combinations and perturbations of external knowledge sources. To enhance efficiency, it incorporates pruning strategies that reduce the search space for counterfactual explanations. Through three challenging use cases, RAGE demonstrates its effectiveness in explaining why LLMs produce specific answers, thereby improving their transparency and interpretability for users.

Another line of work provides ante-hoc explanations to either understand the decision-making process of the generator or enhance the generation performance after incorporating them into the model forward process. In the research of~\cite{luo2024rog}, a novel method called reasoning on graphs (RoG) is proposed to perform such an ante-hoc explanation. The presented framework utilizes knowledge graph relations to generate reasoning paths for the given question, making the reasoning paths interpretable and faithful. Then, an LLM is employed to generate the answer by conducting reasoning on the paths, which is intuitive to understand the decision-making process of the LLM. Through this framework, the generation process can be interpretable and traceable, not only enhancing the explainability in generation but also allowing the RAG system to generate more accurate answers.

\subsection{Dual Enhancement of Explanations and RAG}
In addition to exploring how to explain RAG systems, existing work also investigated the dual enhancement of explanations and RAG. On the one hand, explanations can be integrated to augment RAG systems. On the other hand, RAG systems can also be employed to provide explanations.

As information retrieval plays a critical role in RAG~\cite{cuconasu2024power}, the enhancement of adding explanations during retrieval could also benefit the whole RAG system. The ExaRanker method~\cite{ferraretto2023exaranker} utilizes the explanations as additional labels to train the ranking models in the information retrieval task. Specifically, given the question-passage pair and the label indicating whether the passage can be used to answer the question, an LLM is first employed to generate the explanations of why the question can/cannot be answered by the given passage. With these ground-truth explanations, the ranking model is trained on question-passage pairs to predict not only whether the question can be answered but also the corresponding explanations. By integrating explanations as additional training labels, the ranking model can better understand the relationships between the questions and passages, which could benefit the ranking performance and ease the demand for a large number of training examples. 

Apart from leveraging explanations to augment RAG systems, the reverse relationship—using RAG to improve explanations—is also worth exploring. For example,~\citet{tekkesinoglu2024feature} use RAG in scene-understanding tasks to create explanations through a question-answering approach. For each input with a class label, the model predicts the probability of belonging to that class. To assess the impact of each semantic feature, it generates predictions without specific features, enabling the calculation of feature importance. These outputs, along with features and contrastive cases, contribute to an external knowledge repository for LLMs. This RAG design can thus generate human-friendly and faithful explanations for the prediction model of the scene-understanding tasks. Another work~\cite{hussien2024rag} studying road user behavior also uses RAG to generate explanations. In particular, this work first creates a human-readable document that explains why the road user may/may not have a specific behavior. The document is then processed to form a database, serving as the external knowledge base of the RAG system. Given a tailored prompt and a query derived from the prediction frame, the RAG system will create a detailed explanation of the road user's intention.

\subsection{Explainability Evaluation}

\paragraph{Metrics.} 
As very few works investigate explainability in the context of RAG, we first review the conventional explanation metrics used in explainable artificial intelligence (XAI). \textit{Fidelity} is one commonly used evaluation metric. It measures to which extent the explanation can accurately reflect the decision-making process of the model~\cite{alangari2023exploring}. Mathematically, fidelity is often defined as the proportion of data samples where the predictive model and the explanation produce the same decision, but there are some variations on computing fidelity, such as using Kullback-Leibler divergence between outputs, conditional entropy, and correlation~\cite{nauta2023anecdotal}. Another metric often used is \textit{stability}~\cite{ghorbani2019interpretation,li2020evaluating,plumb2020regularizing}. It measures the consistency of a method in producing similar explanations for similar or closely related inputs~\cite{vilone2021notions}.

To evaluate the explanations of the generator in a RAG framework, \citet{sudhi2024rag} use two key metrics: \textit{significance} and plausibility. In their framework, significance measures whether the explanations capture the core information present in the input. This is quantified using the F1-score and Mean Reciprocal Rank (MRR). On the other hand, \textit{plausibility} is assessed through human evaluation. Annotators identify specific tokens from the input as ground-truth explanations, and the generated explanations are then evaluated against these selected tokens using the F1-score.

\paragraph{Datasets.}
Although the explainability of RAG enhances user trust and transparency in the generated outputs, evaluating the explanations remain challenging due to the lack of standardized datasets specifically designed for this purpose. Currently, researchers often adapt existing QA datasets to test explainability methods. For example, \citet{sudhi2024rag} utilized randomly sampled English and German QA pairs from the validation split of the XQUAD dataset to evaluate their proposed explainer. Similarly, \citet{luo2024rog} demonstrated the self-explanatory capabilities of their RoG method by automatically generating explanations while performing question-answering tasks on two benchmark knowledge graph question-answering (KGQA) datasets: WebQuestionsSP (WebQSP)~\cite{yih2016value} and Complex WebQuestions (CWQ)~\cite{talmor2018web}. These efforts illustrate that, in the absence of standardized benchmarks, evaluation datasets are specific to different tasks and domains, and researchers rely on existing datasets. The development of dedicated datasets for explainability in RAG remains a critical area for future research, offering the potential to advance systematic evaluation and comparison of explainability methods.

\subsection{Future Directions of RAG Explainability}
\paragraph{Integration of Knowledge Graphs and LLMs.} Our work pioneers the integration of knowledge graphs (KGs) and large language models (LLMs) to enhance retrieval faithfulness in Retrieval-Augmented Generation (RAG) systems~\cite{ni2024trustworthyknowledgegraphreasoning}. Unlike existing approaches that rely solely on GNN-based retrieval methods or LLM-based prompting, our hybrid approach leverages the strengths of both. This integration opens opportunities for more robust and versatile solutions. Beyond question-answering tasks, we propose extending retrieval capabilities to semi-structured knowledge bases, providing richer contexts that enhance text generation and personalization. Future research should focus on optimizing the interaction between these components to improve retrieval accuracy and context integration. Additionally, optimizing the structure and accessibility of the knowledge base is essential for achieving better performance in diverse applications.

\paragraph{Explaining Multi-Component RAG Systems.} RAG systems are inherently complex, with multiple interacting components, e.g., retrievers and generators. Explaining the behaviors of these components introduces unique challenges, especially when they are jointly trained, as seen in approaches by \cite{fan2024survey} and \cite{lewis2020retrieval}. In such cases, the retriever and generator often share embeddings or feature representations, making it difficult to disentangle their individual contributions. The advent of novel retrievers, such as those using LLMs as agents for sequential graph traversal \cite{jin2024graph, wang2024knowledge}, further complicates explainability. Traditional differentiable-based explanation methods are often impractical for such systems. Future work should prioritize developing explainability methods that provide insights into both individual components and their interactions within the system. This could include frameworks that combine the outputs of retrievers and generators to analyze their joint contributions to system behavior.

\paragraph{Performance vs. Explainability Trade-Off.} An ongoing debate in explainable AI concerns the trade-off between model accuracy and explainability. Some studies \cite{crook2023revisiting, arrieta2020explainable} argue that optimizing for one often compromises the other, while others contest this notion \cite{bell2022notthatsimple, rudin2022interpretable}, citing a lack of conclusive evidence. In the context of RAG systems, replacing components like retrievers or generators with fully explainable counterparts offers a promising area for exploration. Researchers could investigate the performance implications of such replacements, analyzing whether transparency and interpretability can coexist with high performance. This line of inquiry could yield valuable insights into the relationship between explainability and system effectiveness, particularly when compared to black-box LLM-based approaches.

\paragraph{Evaluation Metrics for Explainability.} Evaluating the quality of explanations in RAG systems is a challenging yet essential area of research. Existing metrics like fidelity and stability are useful but often fail to capture the subjective, context-dependent nature of explainability. Future work should focus on developing domain-specific metrics tailored to user needs and application contexts. For instance, RAG-Ex introduces metrics such as significance and plausibility. Significance evaluates whether explanations align with predefined information, while plausibility assesses whether they reflect annotators' choices. Expanding such metrics to encompass diverse domains and user preferences will provide a more nuanced understanding of explanation quality and its impact on system usability.

\paragraph{Propagation of Misinformation.} The propagation of misinformation remains a critical concern in RAG systems, particularly in scenarios involving multi-step reasoning or complex retrieval processes. Explainability methods that trace the flow of information through RAG components could help identify and mitigate sources of misinformation. For example, analyzing how erroneous retrievals influence generated outputs could inform the design of more robust systems. Future research should investigate how explainability frameworks can incorporate misinformation detection and prevention mechanisms, ensuring that generated responses are not only accurate but also trustworthy.
\section{Accountability of Retrieval Augmented Generation}
Accountability, in the context of AI, refers to the ability to determine whether decisions or outputs align with established procedural and substantive standards and to identify who is responsible when those standards are violated~\cite{doshi2017accountability}. While accountability focuses on policy and standards, we identify one key technical challenge towards accountability: the ability to identify \textit{ownership}. Generative AI poses unique challenges regarding the attribution of 'speech'—that is, determining who should bear responsibility for its outputs. It is crucial to identify and hold the appropriate entities responsible when outputs deviate from procedural or substantive standards. This necessitates robust mechanisms for content traceability, such as watermarking techniques, which can link outputs back to their source models, datasets, or operators.
\textit{Watermarking} thus serves as a critical tool for implementing ownership, enabling the identification of stakeholders responsible for ensuring procedural compliance and addressing any deviations. By providing a transparent means of associating outputs with their origins, watermarking helps bridge the gap between technical and policy-oriented accountability, making it an essential component of the accountability framework for generative AI systems. For the above reasons, we dedicate this section of discussion to the watermarking in LLMs. 

\subsection{Taxonomy of RAG Accountability}
\begin{table}
\centering
\caption{Taxonomy for RAG Accountability}
\label{accountability-taxonomy}
\scriptsize
\begin{tabular}{ccccc}
\toprule
\textsc{Module} & \textsc{Reference} & \textsc{Technique} & \textsc{Type} & \textsc{Year}\\
\midrule

\multirow{3}{*}{\textit{Retrieval}} & Liu et al.~\cite{liu2024survey}
 & Format-based & Text Watermarking & 2024 \\
& Xu et al.~\cite{xu2024hufu} & Embedding-based & Data Watermarking & 2024 \\
& Sun et al.~\cite{sun2022coprotector} & Trigger-based & Data Watermarking & 2022 \\

\midrule

\multirow{4}{*}{\textit{Generation}} 
& Christ et al.~\cite{christ2024undetectable} & Semantic-based & Sentence-level Watermarking & 2024 \\
& Hou et al.~\cite{hou2023semstamp} & Sampling-based & Token-level Watermarking & 2023 \\
& Kirchenbauer et al.~\cite{kirchenbauer2023watermark} & Logit-based & Global Watermarking & 2023 \\
& Yang et al.~\cite{yang2022tracing} & Post-generation & Lexical/Syntactic & 2022 \\
\midrule
\multirow{2}{*}{\shortstack{\textit{Retrieval} \\ + \textit{Generation}}} 
& \multirow{2}{*}{Jovanovic et al.~\cite{jovanovic2024ward}} & \multirow{2}{*}{Integrated Pipeline} & \multirow{2}{*}{Red-Green Token Scheme}  & \multirow{2}{*}{2024} \\ \\

\bottomrule
\end{tabular}
\vspace{-3ex}
\end{table}
\paragraph{Retrieval.}
Accountability in retrieval in this survey refers to embedding watermarks within the sources of retrieval, encompassing both Text Watermarking and Data Watermarking techniques. These approaches aim to safeguard content and data ownership, ensuring accountability and traceability in RAG systems.

\paragraph{Generation.}
Watermarking plays a crucial role in ensuring accountability in the outputs of LLMs, with strate-
gies applicable at different stages of the generation process. Pre-generation watermarking involves
embedding watermarks during the training phase, enabling the model to generate content that inher-
ently contains identifiable markers. In-generation watermarking integrates watermarking algorithms
directly into the text generation process, embedding watermarks as the LLM produces text during
inference. This approach ensures that the watermarks are seamlessly woven into the generated output
in real time. Finally, post-generation watermarking applies text watermarking techniques to LLM-
generated content after the text is produced, embedding markers without requiring modifications
to the generation process itself. Each of these methods provides unique advantages, collectively
strengthening the traceability and authenticity of LLM-generated text. We summarize the RAG accountability taxonomy in Table \ref{accountability-taxonomy}.

\subsection{Accountability in Retrieval} 

 \subsubsection{Text Watermarking} \label{8.2.1} Text Watermarking involves embedding identifiable markers into textual content to protect copyright and authenticate ownership of the author \cite{liu2024survey}.  Format-based watermarking algorithms are commonly employed as they only embed watermarks in the text format without altering the author's contents \cite{liu2024survey}. This includes line or word shifting methods  \cite{brassil1995electronic} and unicode-based approaches \cite{por2012unispach,rizzo2016content, sato2023embarrassingly}. Specifically, the former involves adjusting text lines or words vertically and horizontally, effectively used in image-format texts, while the latter usually involves inserting or replacing Unicode codepoints such as whitespace for watermarking. Beyond the above methods, other innovative techniques have emerged, including variation in text color or font \cite{mir2014copyright} and feature embedding, e.g., bookmarks or variables \cite{iqbal2019robust}. While these approaches signify the promise of format-based watermarking in preserving text copyright, we should also be wary of their potential vulnerability to adversary attacks such as removal using canonicalization \cite{boucher2022bad} and watermark forgery due to the detectable pattern in watermarked text formats \cite{liu2024survey}.

\subsubsection{Data Watermarking} \label{8.2.2}
Data Watermarking addresses the increasing need to protect datasets used during the training of machine learning models, ensuring proper attribution and preventing unauthorized usage. A key technique in this area is backdoor watermarking, which embeds ownership information directly into the trained model by introducing trigger—specific input modifications that prompt unique, identifiable behaviors in the model.  

\textit{Trigger-based watermarking} is widely employed due to its flexibility and effectiveness as a type of backdoor watermarking. These triggers can take various forms, including word- or sentence-level modifications~\cite{sun2022coprotector}, semantically invariant transformations in code~\cite{sun2023codemark}, or distinctive input formats designed to be recognizable~\cite{xu2024hufu}. Embedding such triggers ensures that ownership can be verified through the model's behavior, even if the dataset itself is no longer accessible.

While trigger-based watermarking is robust, its effectiveness depends on careful design to ensure triggers remain inconspicuous yet detectable. Additionally, as models become more complex and versatile, challenges such as ensuring trigger persistence and avoiding unintended activations must be addressed. As the field evolves, continued innovation in embedding mechanisms and detection strategies will be essential for robust and scalable dataset protection.

\subsection{Accountability in Generation}

\subsubsection{Pre-generation Watermarking}
Pre-generation watermarking involves embedding watermarks during the training phase of LLMs, creating inherent markers within the model's outputs. This approach can be categorized into trigger-based watermarks and global watermarks.

Trigger-based watermarking has been described in detail in Section \ref{8.2.2}. As a localized method, trigger-based watermarking relies on specific inputs to reveal ownership, which minimizes its impact on regular outputs but may limit its detection capabilities in broad use cases.

On the other hand, global watermarking ensures pervasive traceability across outputs but requires careful design to balance robustness with imperceptibility, ensuring the watermark does not degrade quality or usability of the generated content. \textit{Global watermarking} embeds markers in all generated outputs, enabling consistent content tracking without the need for specific triggers. This approach integrates watermarking directly into the model's parameters, with methods including sampling-based or logit-based watermark distillation~\cite{gu2023learnability} and reinforcement learning with feedback from watermark detectors~\cite{learninglearning}. Global watermarking's broader applicability makes it an attractive option for large-scale deployment, particularly in scenarios where consistent tracking of all outputs is critical.

Together, pre-generation watermarking provides a proactive means of embedding accountability into LLMs. Future research should focus on hybrid approaches that combine the specificity of trigger-based watermarking with the universality of global watermarking, enabling robust, scalable solutions that balance protection and practicality.

\subsubsection{In-generation Watermarking} 

In-generation watermarking directly embeds watermarks as the LLM produces text in the inference time. Unlike pre-generation watermarking, which embeds markers into the model during training, in-generation watermarking dynamically modifies the generated outputs, providing flexibility and adaptability without altering the model parameters. This can be classified into two main methods: watermarking in logit generation and watermarking in token sampling.

 \textit{Watermarking in logit generation} involves modifying the logits during inference. This approach is both versatile and cost-effective, as it avoids the need for model retraining. Typically, KGW \cite{kirchenbauer2023watermark} partitions the vocabulary into distinct categories, such as red and green token lists, using a hash function that depends on the preceding token. The watermark is embedded by biasing the selection of tokens from one category (e.g., green tokens) during text generation. The detection of KGW involves analyzing the proportion of green tokens in the output and computing a z-score to determine whether the text is watermarked, reported with a high detection performance.  However, its performance can degrade in low-entropy contexts, such as code generation, where token probabilities are unevenly distributed. Optimization techniques, such as entropy-based weighting \cite{lu2024entropy} and sliding window methods \cite{kirchenbauer2023reliability}, have been proposed to enhance detectability and robustness in these challenging scenarios.

In addition to modifying logits, in-generation watermarking can be achieved during token sampling on both the token level and sentence level. Token-level sampling introduces watermarks by biasing the random seed or pseudo-random number generator guiding token selection. For instance, Christ et al. \cite{christ2024undetectable} proposed using a fixed random sequence to verify watermarked outputs by aligning tokens with pre-determined patterns. While effective, this method faces challenges related to robustness against text edits. To address these limitations, sentence-level sampling watermarking focuses on semantic-level modifications. Algorithms like SemStamp \cite{hou2023semstamp} partition the semantic space into watermarked and non-watermarked regions, ensuring that entire sentences maintain watermark integrity even after semantic-preserving modifications. This approach enhances robustness against text editing while enabling more meaningful watermark detection at a higher granularity.

In-generation watermarking offers significant advantages, such as its flexibility and adaptability during inference. However, challenges remain, including mitigating the impact on text quality, enhancing robustness against removal attacks, and achieving public verifiability. Techniques such as fine-grained vocabulary partitioning \cite{fernandez2023three} and semantic-aware watermarking \cite{fu2024watermarking} show promise in addressing these issues.
Future research should explore hybrid methods that combine the strengths of logits-based and sampling-based approaches, providing both robustness and adaptability. Additionally, developing standards for evaluating the effectiveness and detectability of in-generation watermarks across diverse applications will be critical for their broader adoption in real-world scenarios.

\subsubsection{Post-generation Watermarking} 
Post-generation watermarking involves embedding watermarks into already-generated text, which can be categorized into four primary methods: Format-based watermarking, lexical-based, syntactic-based, and generation-based watermarking\cite{liu2024survey}.  

Format-based watermarking has already been discussed in  Section \ref{8.2.1}  which displays vulnerability to adversary attacks due to detectable patterns. To address this,  Lexical-based watermarking advances by word substitution without altering the original textual semantics. Techniques often involve synonyms or contextually appropriate replacements. Early methods, such as those using WordNet or Word2Vec \cite{topkara2006hiding,fellbaum1998wordnet, mitchell2023detectgpt}, were limited by their lack of context awareness, which could compromise text quality. Recent advances, like BERT-based infill models\cite{yang2022tracing, yoo-etal-2023-robust}, have improved context sensitivity, enabling more robust and semantically coherent watermarking. These methods enhance resilience against watermarking removal like reformatting, significantly enhancing robustness.

\textit{Syntactic-based watermarking} modifies the grammatical structure of sentences to embed watermarks. This involves transformations such as adjunct movement, clefting, and passivization, \cite{atallah2001natural}, later expanded with activization and topicalization \cite{topkara2006words}. While effective in embedding watermarks, these methods are often language-dependent and may require customization to adhere to grammatical rules. Excessive syntactic changes can also disrupt the original style and fluency of the text \cite{liu2024survey}.  

\textit{Generation-based watermarking} leverages advanced neural network models to directly generate watermarked text from the original content and a watermark message. These approaches, such as AWT \cite{abdelnabi2021adversarial} and REMARK-LLM \cite{zhang2024remark}, utilize transformer-based architectures to embed high-capacity watermarks while maintaining the quality and naturalness of the text. Techniques like WATERFALL \cite{lau2024waterfall} further enhance fluency by using LLMs for paraphrasing, ensuring seamless integration of watermarks. This method offers high detectability, robustness, and scalability, making it a promising solution for embedding watermarks in LLM-generated content.  

\subsection{Accountability in RAG Systems}

While watermarking techniques for retrieval and generation have been extensively studied, most existing research addresses these stages independently. Few approaches tackle the unique challenges of watermarking in RAG systems, which require seamless integration across both retrieval and generation processes. WARD (Watermarking for RAG Dataset Inference) \cite{jovanovic2024ward} stands out as a pioneering method that integrates watermarking across both stages, distinguishing it from traditional approaches that focus solely on one.

WARD bridges the gap between retrieval and generation watermarking by embedding imperceptible signals into datasets during their creation or integration. This ensures traceability throughout both the retrieval and generation stages. Traditional watermarking techniques either track dataset provenance during retrieval or trace outputs during generation but fail to provide a unified solution for RAG systems. WARD's approach ensures that these embedded signals remain detectable even after transformations common in RAG workflows, such as paraphrasing or reformatting. By modifying token probabilities to embed watermarks, WARD offers a robust and unified solution for dataset ownership attribution that is resistant to obfuscation.

A key strength of WARD is its ability to handle complex scenarios that challenge traditional methods, such as fact redundancy where multiple documents contain overlapping or similar information. Watermarking approaches limited to the retrieval stage may fail to detect dataset usage after generative transformations, while those focusing only on the generation stage may overlook the contributions of source datasets. WARD's innovative red-green token-based scheme generates statistically significant signals that persist throughout the entire RAG pipeline. This enables data owners to confidently identify their datasets even when content is blended with other sources or altered during generation.

Furthermore, WARD provides statistical guarantees for watermark detection, significantly reducing false positives and negatives. Techniques such as aggregated queries enhance reliability without compromising computational efficiency, making WARD both scalable and practical for real-world applications. Its ability to handle large-scale datasets and the inherent complexities of RAG systems further sets it apart from traditional watermarking approaches that consider retrieval and generation stages independently.

Overall, WARD represents a significant advancement in watermarking for RAG systems by effectively integrating techniques across both retrieval and generation stages. Its comprehensive approach not only ensures robust dataset ownership attribution but also addresses the limitations of traditional methods that treat these stages separately. As RAG systems become increasingly prevalent, solutions like WARD are essential for protecting intellectual property and ensuring ethical data usage. Future developments may build upon WARD's framework to enhance watermark resilience and adapt to evolving data transformation techniques.

\subsection{Accountability Evaluation}
\subsubsection{Metrics}

Evaluating watermarking techniques for large language models involves a comprehensive framework of metrics, focusing on detectability, quality impact, output performance, output diversity, and robustness \cite{liu2024survey}. These metrics ensure the effectiveness of watermarking systems while minimizing quality degradation and maximizing resilience against attacks.

\textit{Detectability} is a fundamental metric that evaluates the ability to identify the presence of a watermark in text. For zero-bit watermarking, the focus is on determining whether a watermark exists, without recovering specific information. This is typically achieved using statistical methods such as z-scores or p-values, with careful calibration to minimize false positives, as they can misclassify human-generated text. On the other hand, multi-bit watermarking involves extracting encoded information, which is assessed using metrics like Bit Error Rate (BER) \cite{yoo-etal-2023-robust} and bit accuracy\cite{yoo-etal-2024-advancing}. Additionally, watermark size \cite{perkins2023academic}, which refers to the text length required for reliable detection, is critical, with longer texts generally improving detectability at the cost of applicability in shorter content.

Watermarking techniques must preserve the quality of the generated text, ensuring that the output remains coherent and natural. Quality metrics are divided into comparative and single-text methods. \textit{Comparative metrics} evaluate differences between watermarked and non-watermarked text, using surface-level measures like BLEU \cite{10.3115/1073083.1073135} and Meteor \cite{alkawaz2016concise} or semantic-level measures like Semantic Score and Entailment Score, which leverage embeddings to capture deeper relationships. \textit{Single-text} metrics, like Perplexity (PPL), focus on the coherence of the watermarked text independently. Lower PPL indicates higher fluency and coherence, while human evaluation remains the gold standard for quality assessment.

\begin{table}[t] 
\centering
\scriptsize
\caption{Watermarking Datasets and Metrics} \label{tab-acc}
\begin{tabular}{@{}cccc@{}}
\toprule
\multirow{2}{*}{\textbf{Dataset}} & \multicolumn{3}{c}{\textbf{Metric}}         \\  
                                  & Detectability & Quality Impact & Robustness \\ \cmidrule(r){1-4}
WaterBench  \cite{tu-etal-2024-waterbench}                      &                \Checkmark &                &            \\
WaterJudge   \cite{molenda2024waterjudge}                     &                \Checkmark &                 \Checkmark &            \\
Mark My Words  \cite{piet2023mark}                   &                \Checkmark &                &             \Checkmark \\
MarkLLM    \cite{pan2024markllm}                       &                \Checkmark &                 \Checkmark &             \Checkmark \\ \bottomrule
\end{tabular}
\end{table}

Output performance metrics assess whether the capabilities of watermarked LLMs remain intact across downstream tasks. For text completion, metrics like PPL \cite{yoo-etal-2023-robust}, GPT-4-based scoring \cite{cryptoeprint:2023/1661}, and semantic similarity measures are used to ensure the generated text aligns with prompts. Code generation tasks require precise evaluation using metrics such as CodeBLEU \cite{guan-etal-2024-codeip}, which captures lexical and semantic accuracy, and Edit Sim \cite{tu-etal-2024-waterbench}, which measures the similarity between reference and generated code. Other downstream tasks, including machine translation, summarization, and question answering, are evaluated using standard metrics like BLEU, ROUGE \cite{lin2004rouge}, and task-specific accuracy measures to ensure watermarking does not degrade model utility.

Output \textit{diversity metrics} are essential to evaluate the potential restriction in creativity or variability introduced by watermarking. Metrics like Seq-Rep-N \cite{gu2024on} measure lexical diversity by calculating the ratio of unique n-grams to total n-grams in text. Log Diversity \cite{kirchenbauer2024on} builds on this by quantifying the diversity logarithmically across different n-grams. Additionally, entropy-based metrics such as Ent-3 \cite{hou-etal-2024-semstamp} and Sem-Ent \cite{han-etal-2022-measuring} provide insights into lexical and semantic diversity, respectively. Higher entropy scores indicate greater diversity, ensuring that the watermarking process does not overly constrain model outputs.

Robustness evaluates the ability of watermarked systems to resist adversarial attacks, which can be classified into untargeted attacks and targeted attacks. For details, see Section \ref{sec:robustness} Robustness.

\subsubsection{Datasets}

According to various metrics discussed above, several benchmarks and toolkits have been developed to standardize the evaluation of text watermarking techniques in large language models (LLMs), including WaterBench \cite{tu-etal-2024-waterbench}, WaterJudge \cite{molenda2024waterjudge}, Mark My Words \cite{piet2023mark}, and MarkLLM \cite{pan2024markllm}. The details about each dataset can be viewed in Table \ref{tab-acc}.

\subsection{Future Direction of RAG Accountability}

\paragraph{Unifying Retrieval and Generation Watermarking.} 

Currently, retrieval and generation watermarking techniques operate independently within RAG systems, leaving gaps in overall accountability. Future research should develop unified frameworks that seamlessly integrate both methods. By embedding traceability throughout the entire RAG pipeline, these frameworks would enhance intellectual property protection, ensure responsible attribution, and maintain data and model integrity in complex AI systems where retrieval and generation increasingly overlap.

\paragraph{Dynamic Watermarking for Adaptive AI.}
As generative AI systems evolve to adapt to real-time inputs and changing user contexts, static watermarking becomes insufficient. Developing dynamic watermarking techniques that adjust to system updates, counter adversarial attacks, and respond to shifts in model behavior is crucial. These adaptive methods would enhance robustness and maintain traceability in RAG architectures, supporting accountability in ever-changing AI environments.

\paragraph{Governance and Ethical Integration.}

Besides technical innovation, alignment with legal and ethical frameworks is essential. Future efforts should foster collaborations among technologists, policymakers, and ethicists to establish governance models that incorporate watermarking as a foundational tool for AI accountability and intellectual property protection. Such interdisciplinary work would ensure that RAG systems adhere to global ethical standards and legal requirements.

\section{Applications}
In the preceding sections, we delved into the six dimensions of Trustworthy RAG systems, outlining their foundational principles and challenges. Building on this foundation, we now extend the discussion to their practical applications across high-stakes domains. Specifically, we focus on Healthcare, Legal, and Education. For each domain, we provide a summary of the current landscape and highlight recent advancements. Additionally, we identify domain-specific open problems that present opportunities for further research and development.

\subsection{Healthcare}
\subsubsection{RAG Use Cases in Healthcare}

\paragraph{Clinical Decision Support.}
One of the primary healthcare applications of LLMs is in Clinical Decision Support (CDS) for medical professionals~\cite{LEVRA2025113}. These systems enhance the decision-making process by providing access to diagnostic guidelines, treatment plans, and patient history for medical professionals. Although the majority of clinical models leverage fine-tuning for knowledge injection, recent results have shown that by integrating the retrieval of relevant medical documents and personalized patient data, RAG systems can improve the performance of the model to make more informed and accurate decisions~\cite{xiong2024benchmarkingretrievalaugmentedgenerationmedicine}. 

\paragraph{Patient Communication.}
While CDS systems are geared toward assisting medical professionals, RAG systems can also play a vital role in patient-facing communication. There are mainly two categories of patient-facing communication: question answering (QA) and dialogue support~\cite{HE2025102963}. The Centers for Disease Control and Prevention (CDC) has shown that 58\% of US adults have used the internet to search for medical information~\cite{cdc2023internet}. This widespread reliance on online health information underscores the need for personalized and reliable medical guidance. RAG systems, by integrating structured retrieval with generative capabilities, can enhance the accuracy and relevance of patient interactions. For instance, a RAG-based chatbot might help patients assess symptoms and recommend next steps, such as scheduling a medical consultation or seeking immediate care. 

\paragraph{Knowledge Discovery.}
AI has been extensively used in various medical knowledge discovery tasks such as drug discovery. Traditional works prioritize graph-based approaches for analyzing molecular structures and drug-drug interactions, leveraging methods such as GNN~\cite{Rohani2019, Rabeah2022}. Recently, researchers started to look at LLM for science and explore the application of LLMs in medical-related fields, particularly for accelerating drug discovery and development through improved information processing and hypothesis generation~\cite{Pal2023}. RAG systems can significantly accelerate the research process by retrieving the latest research findings, publications, and clinical trial data~\cite{Yang2025}. For example, research has shown that RAG can greatly enhance the prediction on Nephrology~\cite{Miao2024RAG}. By synthesizing large volumes of medical literature into a retrieval database, these systems provide valuable insights that can inform future studies, improve hypotheses, and guide the direction of medical innovation~\cite{Roy2024}.

\paragraph{Precharting.}
Precharting is a crucial healthcare application that involves a physician reviewing a patient's medical records before a visit~\cite{Bowman2021EMR}. Studies have demonstrated that LLMs encode extensive medical knowledge~\cite{Singhal2023LLM}, while retrieval-augmented generation (RAG) can enhance personalization and contextual relevance in clinical settings~\cite{Yang2025}. Although the integration of RAG into precharting remains largely unexplored, these observations suggest its potential to transform the process by improving efficiency and reducing load of healthcare professionals. Future research could further investigate how RAG can optimize precharting workflows and enhance patient care.

\subsubsection{Trustworthiness Challenges in Healthcare}
\paragraph{Reliability.}
 Previous research have identified reliability as one of the main challenge to healthcare applications~\cite{HE2025102963}. Healthcare applications demand an exceptionally high level of reliability due to the high-stakes impact of decisions made using RAG outputs. Uncertainty quantification has been extensively used in various healthcare applications~\cite{SEONI2023107441}, but its use in LLM-based applications remains largely unexplored. Applications such as CDS and Patient Communication should include uncertainty quantification measures to enable end-users to make informed decisions by understanding the confidence levels of the outputs. Furthermore, these systems require robustness to ensure consistent performance under challenging or unexpected input. Current research often focuses on general question answering~\cite{ni2024trustworthyknowledgegraphreasoning,luo2024rog, sun2023thinkongraph}; however, we advocate for increased focus on domain-specific and individual difference challenges in medical knowledge. Tailoring RAG systems to the requirements of healthcare can enhance their reliability and effectiveness in real-world applications.

\paragraph{Privacy.}
Another critical challenge in healthcare RAG applications is maintaining the privacy of sensitive medical data. The personal nature of medical information makes it imperative to safeguard against data breaches and misuse. Zeng et al.~\cite{privacy_rag_2024} highlight the risks of personal identification information leakage in medical data, representing an initial effort to address this issue. Since applications such as precharting heavily rely on the retrieval of personal medical records, ensuring the privacy of the data is imperative given the high stakes~\cite{Bowman2021EMR}. Future research should extend these efforts to more complex, real-world scenarios where multiple stakeholders, data sources, and regulatory frameworks intersect. Developing privacy-preserving mechanisms tailored to healthcare RAG systems will be crucial to fostering trust and ensuring compliance with legal and ethical standards.

\paragraph{Others.}
While Reliability and Privacy are particularly critical in healthcare applications, the other dimensions of trustworthiness also play significant roles. For instance, healthcare RAG systems must be safe and secure, i.e., resilient to adversarial attacks, as malicious actors could manipulate critical outputs, potentially leading to harmful consequences for patients~\cite{GhaffariLaleh2022Adversarial}. Fairness is equally essential, as healthcare is a fundamental right, and biases in RAG outputs could disproportionately disadvantage certain populations or exacerbate health disparities~\cite{HE2025102963}. Additionally, Explainability and Accountability are crucial to building trust with both medical professionals and patients. Explainable systems allow users to understand the rationale behind recommendations, while accountability mechanisms ensure that errors or unintended consequences can be traced and addressed effectively. Together, these dimensions create a comprehensive framework for trustworthy healthcare RAG systems.

\subsection{Law}
\subsubsection{RAG Use Cases in Law}
\paragraph{Legal Question Answering.}
One of the key applications of RAG systems in the legal field is a legal question answering (LQA)~\cite{lai2023largelanguagemodelslaw, chen2024survey}. LQA systems aim to provide answers to queries related to law, cases, and theoretical analysis. Due to the idiosyncratic nature of the laws in different jurisdictions, current methods focus on fine-tuning existing language models on specific laws or jurisdictions~\cite{ahmad-etal-2020-policyqa, mansouri2023falqufindinganswerslegal, abdallah2023exploring}. These systems streamline the process of finding precedents, statutes, and relevant case law, enabling legal professionals to build stronger arguments and make informed decisions. Recent research has shown that RAG can also help provide useful context for LQA~\cite{Wiratunga2023CBR} besides fine-tuning. Given the highly specialized nature of law, future research should focus on effectively integrating context to LQA through various methods (Fine-tuning, RAG, etc.). 

\paragraph{Legal Document Summarization.}
Legal Document Summarization (LDS) has become an increasingly popular use case of large language models (LLMs) in the legal community~\cite{Anthropic_Legal_Summarization, chen2024survey}. Legal documents are often long and complex, making manual summarization time-consuming and error-prone. LDS assists legal professionals and researchers by enhancing efficiency in legal analysis. Commercial models such as Claude have incorporated rule-based summarization techniques alongside state-of-the-art LLMs to improve the quality and relevance of summaries. However, the integration of RAG into LDS remains underexplored. By incorporating RAG into LDS systems, legal summaries can better capture relevant case law, statutes, and contextual references, leading to improved factual consistency and enhanced robustness~\cite{liu2024robustretrievalbasedsummarization}. 

\paragraph{Legal Judgment Prediction.}
Legal Judgment Prediction (LJP) aims to forecast court rulings based on case fact descriptions~\cite{chen2024survey}. It plays a crucial role in legal decision support, assisting judges, lawyers and clients in analyzing case outcomes. Early research formulated LJP as a classification task using traditional machine learning models~\cite{cui2022surveylegaljudgmentprediction}. However, to better reflect real-world judicial reasoning, recent studies have integrated external legal databases and RAG techniques to incorporate precedent cases and legal statutes~\cite{wu-etal-2023-precedent}, enhancing the interpretability and reliability of predictions. 

\subsubsection{Trustworthiness Challenges in Legal Applications}
\paragraph{Fairness.}
Fairness is a critical concern in legal RAG applications, as biases in data or algorithms can lead to unjust or discriminatory outcomes. For instance, biases in training data may disproportionately affect marginalized groups by retrieving prejudiced legal precedents or overlooking relevant cases. Existing research has shown that LLMs and RAG systems are susceptible to both explicit and implicit biases~\cite{shrestha2024fairrag, wan2023genderbiases}. While initial methods have been proposed to mitigate these biases~\cite{10.1162/coli_a_00524}, dedicated research in legal contexts remains limited. Addressing fairness in legal RAG systems requires not only debiasing training data but also designing algorithms that promote equity in document retrieval and output generation. Future research should explore domain-specific challenges to ensure just and transparent legal RAG systems.

\paragraph{Explainability.}
Explainability is essential in legal RAG applications due to the complexity and high-stakes nature of legal decision-making~\cite{cui2022surveylegaljudgmentprediction, chen2024survey}. Legal professionals require transparent insights into how these systems retrieve and synthesize information. For instance, when generating judicial opinions, it is crucial to trace the origins of retrieved precedents and legal arguments to assess their relevance, credibility, and potential biases. While existing research has explored explainability in general LLMs, recent studies have specifically examined interpretable methods for long-form legal question answering~\cite{louis2023interpretablelongformlegalquestion}. Future work should focus on developing domain-specific explainability frameworks that ensure legal RAG systems provide justifications aligned with legal reasoning.

\paragraph{Others.}
In addition to Fairness and Explainability, other dimensions of trustworthiness play a vital role in legal RAG applications. Reliability needs to be ensured as the output of the legal applications needs to be factually grounded with reliable uncertainty and robustness estimation. Privacy is paramount when handling sensitive client information or confidential case details, necessitating the development of privacy-preserving mechanisms compliant with legal and ethical standards. Furthermore, similar to the healthcare applications, accountability are essential to fostering trust among legal professionals and clients. Accountability mechanisms ensure that errors, biases, or unintended consequences can be identified and addressed. Together, these dimensions establish a comprehensive foundation for trustworthy RAG systems in legal applications.

\subsection{Education}
\subsubsection{RAG Use Cases in Education}
\paragraph{Personalized Learning.}
LLMs have demonstrated significant potential in education~\cite{wang2024largelanguagemodelseducation}, with personalized learning being one of their key applications. Recent advancements in educational research highlight the effectiveness of LLMs in learning path planning~\cite{ng2024educationalpersonalizedlearningpath}. However, existing approaches often lack dynamic retrieval mechanisms to adapt to different student needs and contextual knowledge gaps. Integrating RAG can enhance personalized learning by incorporating up-to-date information, ensuring more adaptive and tailored learning experiences. This approach has the potential to improve student engagement, optimize learning trajectories, and foster deeper comprehension.

\paragraph{Student Support.}
RAG systems can assist students by providing real-time answers to academic queries and offering guidance on educational content~\cite{wang2024largelanguagemodelseducation, dakshit2024facultyperspectivespotentialrag}. For example, a RAG-powered chatbot could help students understand complex concepts that might not be necessarily present in the training material. By addressing common questions and providing actionable insights, these systems can reduce the burden on teachers and counselors while providing valuable assistance to the students. 

\paragraph{Teacher Support.}
Another valuable application of RAG systems is assisting teachers in classroom instruction by integrating external databases to provide up-to-date knowledge and resources. This can help educators generate lesson materials and answer student queries based on real-time information. Current research has shown promising preliminary results in leveraging LLMs for classroom support using reddit as a data source~\cite{mullins2024enhancingclassroomteachingllms}. However, there remains a gap in developing high-quality, domain-specific datasets and more sophisticated models tailored for real-world classroom settings. Future research should focus on refining dataset curation and improving model adaptability to better integrate RAG into educational practices.

\subsubsection{Trustworthiness Challenges in Educational Applications}
\paragraph{Fairness.}
Recent research has highlighted that LLMs often exhibit explicit and implicit biases~\cite{wan2023genderbiases, sheng2021nice, sheng2021societal}. In the context of educational applications, ensuring fairness is crucial to provide all students with equal access to learning opportunities and unbiased content. For example, a RAG-based learning system might retrieve or synthesize information influenced by biases present in its training data, potentially reinforcing stereotypes or disadvantaging students from underrepresented groups. Furthermore, if the system inadvertently retrieves or generates content based on protected attributes such as race, gender, or socioeconomic status, it could lead to unequal treatment or negative educational outcomes. 

\paragraph{Safety.}
As shown in our previous discussion, studies have demonstrated that RAG systems are vulnerable to various adversarial attacks~\cite{deng2024pandorajailbreakgptsretrieval, xue2024badrag, wei2024jailbroken}. These vulnerabilities present significant safety concerns in educational settings. For instance, a student using a RAG-based learning platform could be exposed to harmful or misleading content if a malicious actor successfully jailbreaks the system to bypass safety filters. Such breaches could lead to the dissemination of inappropriate or even dangerous materials, undermining the utility of the system and harming the minors. To mitigate these risks, robust adversarial defense mechanisms tailored to the application are essential to ensure that educational RAG applications remain secure and reliable for learners.

\paragraph{Others.}
Beyond Fairness and Safety, other dimensions of trustworthiness are equally critical for educational applications of RAG systems. First, Reliability must be ensured to emphasize the accuracy and consistency of retrieved or synthesized content. Next, Privacy is a vital consideration, as educational platforms often handle sensitive personal information such as student performance, learning behaviors, and demographic data. A failure to protect privacy not only undermines trust but could also expose students to risks such as data breaches or identity theft. Lastly, Robustness is crucial for maintaining system trustworthiness in dynamic educational environments where students and learning objectives might constantly change. Educational RAG systems must adapt to diverse user queries, varying levels of prior knowledge, and potential ambiguities in input while ensuring stable performance.

\section{Conclusion}
In this survey, we provide a comprehensive review of RAG systems through the lens of trustworthiness, focusing on six critical aspects: reliability, privacy, safety, fairness, explainability, and accountability. This work addresses the pressing need for a unified perspective in the field, bridging the gap in understanding and systematically categorizing the challenges and solutions in developing trustworthy RAG systems.
For each trustworthiness aspect, we introduce definitions and key concepts to establish an understanding of the topics. We also offer a structured taxonomy to help researchers navigate the diverse approaches in the specific trustworthy aspect. Beyond methodological summary, we highlight the evaluation protocols commonly used for trustworthy RAG systems, including the specific datasets and metrics. By including these discussions, we aim to facilitate the development of benchmarks tailored to trustworthiness challenges. 

Finally, we provide an in-depth discussion of future research directions for each aspect of trustworthiness, including promising directions within individual aspects and potential synergies across multiple areas. By addressing these directions, we hope to inspire innovative approaches that enhance the overall trustworthiness of RAG systems and drive their broader adoption in critical applications. This survey not only serves as a comprehensive roadmap for researchers aiming to advance trustworthy RAG systems but also underscores the importance of addressing these challenges to ensure the safe and ethical deployment of AI technologies.




\bibliographystyle{ACM-Reference-Format}
\bibliography{main}

\appendix

\end{document}